\documentclass{article}

\usepackage[final]{neurips_2022_arxiv}

\usepackage{natbib}
\usepackage[utf8]{inputenc} 
\usepackage[T1]{fontenc}    
\usepackage{hyperref}       
\usepackage{url}            
\usepackage{booktabs}       
\usepackage{amsfonts}       
\usepackage{nicefrac}       
\usepackage{microtype}      
\usepackage{amsthm}
\usepackage{amsmath}
\usepackage{parskip}
\usepackage{amsthm}
\usepackage{wrapfig}
\usepackage{subcaption}

\usepackage{graphicx}
\usepackage{braket}
\usepackage{microtype}
\usepackage{graphicx}
\usepackage{booktabs} 
\usepackage{maths}

\everypar{\looseness=-1}

\usepackage{xcolor}
\hypersetup{
	colorlinks,
	linkcolor={red!40!gray},
	citecolor={blue!40!gray},
	urlcolor={blue!70!gray}
}

\usepackage[capitalize,noabbrev]{cleveref}

\theoremstyle{plain}
\newtheorem{theorem}{Theorem}

\theoremstyle{proposition}
\newtheorem{proposition}[theorem]{Proposition}
\theoremstyle{lemma}
\newtheorem{lemma}{Lemma}
\theoremstyle{corollary}
\newtheorem{corollary}[theorem]{Corollary}
\theoremstyle{definition}
\newtheorem{definition}{Definition}
\theoremstyle{assumption}

\theoremstyle{remark}

\DeclareMathOperator*{\argmax}{arg\,max}
\allowdisplaybreaks

\title{Experimental Design for Linear Functionals in Reproducing Kernel Hilbert Spaces}

\author{%
  Mojm\'ir Mutn\'y \\ ETH Z\"urich \\ \texttt{mojmir.mutny@inf.ethz.ch} \And Andreas Krause \\ ETH Z\"urich \\ \texttt{krausea.ethz.ch}
}

\begin{document}

\maketitle

\begin{abstract}
	\looseness -1 Optimal experimental design seeks to determine the most informative allocation of experiments  to infer an unknown statistical quantity. In this work, we investigate the optimal design of experiments for {\em estimation of linear functionals in reproducing kernel Hilbert spaces (RKHSs)}. This problem has been extensively studied in the linear regression setting under an estimability condition, which allows estimating parameters without bias. We generalize this framework to RKHSs, and allow for the linear functional to be only approximately inferred, i.e., with a fixed bias. This scenario captures many important modern applications, such as estimation of gradient maps, integrals, and solutions to differential equations. We provide algorithms for constructing bias-aware designs for linear functionals. We derive non-asymptotic confidence sets for fixed and adaptive designs under sub-Gaussian noise, enabling us to certify estimation with bounded error with high probability.
\end{abstract}

\section{Introduction}
Optimal Experimental Design (OED) aims to determine data collection schemes -- \emph{designs} -- to efficiently estimate unknown quantities of interest given limited resources \citep{Chaloner1995}. As common, we model experiments via an {\em oracle} that yields a (noisy) response to a given input.  A design is usually either an (adaptive) policy for querying the oracle, or a (nonadaptive) fixed allocation of query budget to different oracle inputs. OED has a rich history and close relations to the field of bandits \citep{Szepesvari2019} and  active learning \citep{Settles2009}.

We consider the regression setting, where observations at a fixed input $x \in \mX$ can be obtained via
the noisy oracle by:
\begin{equation}\label{eq:model}
	y = \theta^\top\Phi(x) +  \epsilon ~ \text{where} ~ x \in \mX
\end{equation}
and $\cdot^\top\cdot$ denotes the inner product in a Hilbert space $\mH_\kappa$, $\epsilon$ is independent, sub-Gaussian noise with \emph{known} variance proxy $\sigma^2$, and $\theta$ is a bounded element from a separable reproducing kernel Hilbert space $\mH_\kappa$ from kernel $\kappa$, with a bound $\theta^\top \mV_0\theta \leq \lambda^{-1}$, where $\mV_0:\mH_k \rightarrow \mH_k$ is a positive definite operator. \footnote{In most cases, but not all, $\mV_0$ is chosen to be the identity.} Depending on whether $\lambda$ is known or unknown, we will propose different estimators. The norm constraint (under $\mV_0$) can be interpreted as a bound on the total effect and its parameters if the Hilbert space is finite dimensional. For infinite dimensional spaces this assumption can model the additional constraint on regularity. For example, for kernels that induce Sobolev spaces, it explicitly bounds the total squared derivatives of functions -- quantities known to govern complexity of approximation difficulty. The exact bound depends on the specific definition of $\kappa$, but if two spaces are isometrically isomorphic, the difference can be absorbed to $\mV_0$ to induce the desired prior regularity, e.g., the regularity of the derivatives. Here we assume that kernel $\kappa$ and the operator $\mV_0$ are known due to prior analysis and/or first principles modeling of the system. 

In contrast to the classical ED task of estimating  $\theta$  \citep[see][]{Fedorov1997}, we are interested in estimating a {\em projection} of $\theta$. Namely, let $\bC: \mH_\kappa \rightarrow \mR^{p}$ be a \emph{known linear operator}. The map $\bC$ is such that $\bC^\top \bC$ is full rank $p$, where $\bC^\top$ denotes the adjoint. Our goal is to identify an estimate of $\bC\theta$ efficiently, i.e., with low query complexity or with a maximal reduction of uncertainty given a fixed query budget $T$.

The formalism above captures, or occurs as a subroutine in, numerous practical problems. For example, {\em evaluation} at specific target points, {\em integration} and {\em differentiation} are all linear operators, among many other useful linear functionals. Other examples include ordinary or partial differential equations operators, spectral transforms, and stability metrics from control engineering. In Section~\ref{sec:example} we detail several example applications.

A naive approach would be to first obtain an estimate $\hat{\theta}$ of $\theta$, and compute $\bC\hat{\theta}$. However, the appeal of estimating linear functionals {\em directly} is that the number of unknowns of interest may be {\em much lower} than for the original overall unknown element $\theta$ (which might even be infinite-dimensional). Consequently, we would hope that the query complexity of reducing the variance of the estimate $\bC \theta$ scales in the dimension of the range of $\bC$, which is $p$. For example, when focusing on finite-dimensional RKHSs, the operator $\bC:\mR^{m} \rightarrow \mR^p$, where $m$ is the dimension of $\mH_\kappa$, becomes a matrix. In this work, we study the cases where $p<m$, and show that the estimation error can indeed scale with $p$, and the geometry of the query set $\{\Phi(x) | x \in \mX\}$. The estimation {\em bias} plays a central role in this work as for very large $m$ estimating $\bC\theta$ might not be possible up to any precision. The difficulty depends among other things the richness of the query set $\mX$, as we will detail later in Sec. \ref{sec:estimators}.
\vspace{-0.15cm}
\paragraph{Sequential Experiment Design}
Apart from classical experiment design, where we first commit to what set of queries $x$ 
we choose, referred to as a \emph{fixed design}, we also consider sequential design, where the selected queries depend on past observations. Suppose for example, we are gathering data to test whether the null $H_0$: $\theta^\top \Phi(x) \geq 0$ for all $x\in \mX$ or otherwise. We can incrementally gather evidence and check whether the null hypothesis has already been rejected. As our data depends on prior evaluation points it forms an \emph{adaptive design}. In this work, we develop confidence sets for both fixed and \emph{adaptive designs} -- which are of paramount importance in the context of sequential experiment designs, e.g., to define stopping rules of adaptive hypothesis testing problems.

%
%

\vspace{-0.1cm}
\paragraph{Contributions}
\looseness -1 \textbf{A)} We consider objectives for experiment design for linear functionals in general RKHS spaces, which carefully take the bias of the estimator into account.
\textbf{B)} We provide bounds on the query complexity required to reach $\epsilon$ accuracy to estimate linear functionals with high probability.
\textbf{C)} We construct novel non-asymptotic confidence sets for linear estimators of linear functionals of RKHS elements, both for {\em fixed} and {\em adaptive} designs, where queries are {\em independent} of previous noise realizations, and where they are not, respectively.
\textbf{D)} We demonstrate the improved inference error due to specially defined designs and new confidence sets on the problems of learning differential equations, linear bandits, gradient maps estimation, and stability verification of non-linear systems.
%
\vspace{-0.2cm}
\section{Background and Related Work}\label{sec:background}
\paragraph{Linear Estimators} \looseness -1 Let $\mS \subset \mX$ be a finite set of selected evaluations s.t. $|\mS| = n$, potentially repeated. We focus on linear estimators of the form $\bL:\mR^{n} \rightarrow \mR^p$, where $\bL y$ is estimating $\bC \theta$. An \emph{estimator} is understood here as the algorithm to find the random \emph{estimate} $\bL y$. Notice that given $\mS$, $\bL$ is not a random quantity; the randomness rather comes from the realizations $y$. To choose the estimator $\bL$, one classically looks at the second moment of the residuals $\bC\theta - \bL y$,
$\bE(\bL) = \mE[(\bC\theta - \bL y)(\bC\theta - \bL y)^\top]$, where the argument signifies how the random variable $y$ is transformed (noise is averaged)
\begin{equation}\label{eq:residuals}
	\bE(\bL) =  \sigma^2\bL\bL^\top + (\bL\bX - \bC)\theta\theta^\top (\bL\bX - \bC)^\top, \quad \text{where} \quad \bE(\bL) \in \mR^{p \times p},
\end{equation}
and the matrix $\bX$ contains stacked evaluation functionals of the RKHS $\bX_{i,x} = \Phi_i(x)$ for $ x\in \mS \subseteq \mX$. We will then seek a way to transform $y$ via estimator $\bL$ such that the second moment of residuals is minimized in certain sense.
\vspace{-0.1cm}
\paragraph{Importance of Bias and RKHS}
\looseness -1 Classical experimental design studies estimation of $\bC: \mR^{m} \rightarrow \mR^{p}$ where the RKHS is {\em finite} dimensional \citep{Pukelsheim2006} and $\bX\in \mR^{n \times m}$. On top of that, they consider only estimators which are {\em unbiased} for {\em any} $\theta$, in other words, $\bL\bX =\bC$. While, with finite dimensions, this simplification might be reasonable, for infinite dimensionial RKHSs, bias is {\em inevitable} and must be controlled. Consider the case of \emph{estimating the gradient} of a continuous function $f$. We can nearly never learn it up to arbitrary precision from noisy point queries. However, estimation up to a small error is always possible and sufficient. Estimating $\nabla_x f(x)$ from points close to $f(x)$ will incur small bias but larger variance as the change in $f$ compared noise $\epsilon$ is small. Hence balancing these two sources of error is crucial for an informative design in RKHSs.
\vspace{-0.1cm}
\paragraph{Experiment Design: Classical Perspective}
\looseness -1 Consider an $m$ dimensional version of the model in \eqref{eq:model}. Then, among all unbiased estimators, linear least-squares estimators minimize \eqref{eq:residuals} under the L\"owner order due to the famed Gauss-Markov theorem. Unbiasedness in this form is synonymous with {\em estimability}, which means that by repeating the evaluation in $\bX$, arbitrary precision can be reached \citep{Pukelsheim2006}.
The second moment of residuals then becomes $\bW_\dagger^{-1} = \bC \bV^\dagger \bC^\top$, where $\cdot^\dagger$ denotes a generalized pseudo-inverse and $\bV = \bX^\top \bX$. The matrix $\bW$ is often referred to as the \emph{information matrix}. \citet{Gaffke1987} and \citet{Krafft1983} note that estimability implies that $\bW_\dagger$ is non-singular. We relax this condition and  allow the estimator to have a {\em bias}; this means that we cannot in general reduce the error arbitrarily by repeating the measurements. In fact, our extension uses the fact that $\bW_\dagger$ as defined above will not be singular even if estimability condition is not satisfied. We will show that the matrix $\bW_\dagger$ can still play the role of the information matrix.
\vspace{-0.1cm}
\paragraph{Experiment Design: Modern Challenges}
\looseness -1 \citet{Mutny2020} use experimental design to estimate the Hessian of an unknown RKHS function, while \citet{Kirschner2019} use it to estimate the gradient of it for use in Bayesian optimization. Perhaps most related, \citet{Shoham2020} study over-parametrized experimental design for one-shot active deep learning and analyze the bias in connection to experiment design, similarly as in the seminal work of \citet{Bardow2008}. They do not consider bias arising from the limited design space nor do they treat linear functionals.  More broadly, the uncertainty propagation is studied in the Bayesian framework in the field of \emph{probabilistic numerics} for linear and nonlinear operators \citep[see][and citations therein]{Cockayne2017,Owhadi2016}.
\vspace{-0.15cm}
\paragraph{Confidence sets} \looseness -1 Unlike classical statistics, our focus is on non-asymptotic confidence sets on regression estimators. They can be found in, e.g., \citet{Draper2014} and  \citet{Abbasi-Yadkori2011}, for fixed and adaptive designs, respectively. Our goal is to define the confidence sets in the appropriate norm such that their width scales with the dimension of the range of $\bC$, i.e., $p$, and the geometry of $\{\Phi(x)|x \in \mX\}$, but not directly $\dim(\mH_\kappa)$ nor number of data $T$. \citet{Mutny2020} derive \emph{non-asymptotic} confidence sets that scale in $p$, but grow with the number of points $T$ as they do not use the appropriate norm (see Appendix \ref{app:relation-prior}). Similarly, \citet{Khamaru2021} study estimators for adaptively collected data, and propose \emph{asymptotic confidence} sets for their estimators. Without a specific condition, their confidence sets can grow with $T$, however, they consider more general noise distributions apart from sub-Gaussian as we do here.


\section{Estimation and Bias}\label{sec:estimators}
In this section, we motivate the linear estimators and identify \emph{information matrices}. Information matrices are the inverses of second moment matrices of residuals $\bE$ as in Eq. \eqref{eq:residuals}. They are an important object in the analysis of the error of the estimators and their confidence sets. Also, they depend on the evaluations that define the estimator $\bL$, $\bX$. \emph{Maximizing the information matrices} as a function of the chosen observations and their proportions $\bX$ gives rise to \emph{optimal experimental designs.}

Further, we identify quantities that influence the bias of the estimators. We study two estimators: the least-norm estimator (interpolation), when the bound on $\norm{\theta}_{\mV_0}$ is unknown but finite, or the ridge regularized least squares estimator, where $\norm{\theta}_{\mV_0}\leq \lambda^{-1}$ is known. Both estimators are motivated as minimizing the error residuals $\bE$ in trace norm under these two bound assumptions.
\vspace{-0.1cm}
\subsection{Estimators}

\paragraph{Interpolation}
As apparent from Eq.~\eqref{eq:residuals}, without the knowledge of an explicit bound on the norm $\theta$, the worst case over $\theta$ causes the optimal estimator $\bL$ to minimize \emph{only} the trace of the bias. This leads to minmization of $(\bC-\bL \bX)\mV_0^{-1/2}$ in Frobenius norm, leading to the familiar \emph{interpolation estimator},
\begin{equation}\label{eq:interpolator}
	\bC\hat{\theta} := \bL_\dagger y = \bC \mV_0^{-1}\bX^\top\bK^{-1}y, \quad  \text{where}  \quad \bK = \bX\mV_0^{-1}\bX^\top
\end{equation}
The second moment of residuals can then be expressed as
\begin{eqnarray*}\label{eq:covar}
	\bE(\bL_\dagger) \preceq \underbrace{\sigma^2 \bC\mV_0^{-1}\bX^\top \bK^{-2}\bX\mV_0^{-1}\bC^\top}_{\text{variance}} + \underbrace{\frac{1}{\lambda} \bC  \mP_{\bX}  \bC^\top}_{\text{bias}},
\end{eqnarray*}
\vspace{-0.55cm}

where $\mP_{\bX} =  \mV_0^{-1/2}(\mV_0^{-1/2}\bX^\top \bK^{-1} \bX\mV_0^{-1/2} - \mI) \mV_0^{-1/2}$ is a scaled projection matrix. Unlike as in the classical ED treatment, the second moment has two terms: bias and variance. If the span of the scaled projection operator lies in the null space of $\bC$, the bias (classically) vanishes. To control the error of estimation, we need to control both terms. For the special case of interpolation estimator, we will control the second term separately using a \emph{bias condition}, and the variance term will be controlled by the \emph{information matrix} $\bW_{\dagger}$ as we will see in Section \ref{sec:bias}
\begin{equation}\label{eq:info-matrix}
	\bW_\dagger(\bX) = (\bC\mV_0^{-1}\bX^\top \bK^{-2} \bX\mV_0^{-1}\bC^\top)^{-1}.
\end{equation}
\vspace{-0.5cm}
\paragraph{Regularized Regression}
The ridge regularized estimator is motivated by using $\theta^\top\mV_0\theta \leq \lambda^{-1}$, and minimizing then the trace of this upper bound, leading to an estimator,
\begin{equation}\label{eq:ridge}
	\bC\hat{\theta}_{\lambda} := \bL_\lambda y = \bC \mV_0^{-1}\bX^\top(\lambda \sigma^2 \bI + \bK)^{-1}y.
\end{equation}
Like above, we can give an upper bound on the second moment of the residuals. Conveniently, this estimator automatically balances the error due to variance and bias in one term:
$ \bE(\bL_\lambda) \preceq \bC\theta\theta^\top\bC^\top -\bC\mV_0^{-1}\bX^\top(\bI\sigma^2\lambda + \bK)^{-1}\bX \mV_0^{-1} \bC^\top. $
\looseness -1 Using the matrix inversion lemma, we can express the above bound in a more concise form, and subsequently its inverse motivates the definition of the information matrix for the regularized estimator,
\begin{equation}\label{eq:infomatrix-regularized}
	\bW_{\lambda}(\bX) = \sigma^{-2}(\bC(\sigma^2\lambda \mV_0 + \bX^\top\bX)^{-1}\bC^\top)^{-1}.
\end{equation}
\vspace{-0.7cm}
\subsection{Design objectives: Scalarization}\looseness-1 \label{sec:designs}
The information matrices such as in  \eqref{eq:info-matrix} and \eqref{eq:infomatrix-regularized} represent the inverse of estimation error, and the goal of optimal design is to maximize them (hence minimizing the error) with a proper choice of $\bX$. As the L\"owner order is not a total order, we need to resort to some scalarization of the information matrices,  thus we solve $\max_{\bX}f(\bW(\bX))$, where $f$ is the scalarization $\mR^p \rightarrow \mR$. We focus on two common forms of scalarization and refer to them as $E$- and $A$-design \citep{Pukelsheim2006}.
\begin{itemize}
	\item $f_E(\bW) = \lambda_{\text{min}}(\bW)$ -- when constructing non-asymptotic high probability confidence sets;
	\item $f_A(\bW) = 1/\Tr(\bW^{-1})$ -- when minimizing mean squared error of estimation.
\end{itemize}
Other popular criteria include $D$, $V$, and $G$-designs \citep{Chaloner1995}, which we do not consider due to space considerations, but can be equivalently used should the experimenter have a reason for it. If $p = 1$, $\bC$ is an element in $\mH_k$ and the design problem degenerates into a special case known as c-optimality due to \citet{Elfving1952} and, as $\bW\in \mR$, no scalarizations are needed.
\vspace{-0.1cm}
\paragraph{Robust Designs}
The linear functionals are sometimes unknown, or parametrized by an \emph{unknown} parameter $\gamma$ as $\bC_{\gamma}$, where $\gamma$ belongs to a known set $\Gamma$. If we were to construct a design that has low estimation error for the worst case selection of $\gamma$, we can maximize the \emph{information} in the following worst case metric
$\max_{\bX}\inf_{\gamma \in \Gamma} f\left(\left(\bC_\gamma\mV_0^{-1}\bX^\top \bK^{-2} \bX\mV_0^{-1}\bC_\gamma^\top\right)^{-1}\right),$
for the interpolation estimator (and analogously for ridge regression). If the original function $f(\bW)$ is concave, then so is the function defined as the infimum over the compact index set $\Gamma$, which is true for the design criteria considered in this work \citep{Boyd2004}.
\vspace{-0.1cm}

\begin{wrapfigure}[14]{r}{0.5\textwidth}
\vspace{-0.60cm}
\includegraphics[width=0.5\textwidth]{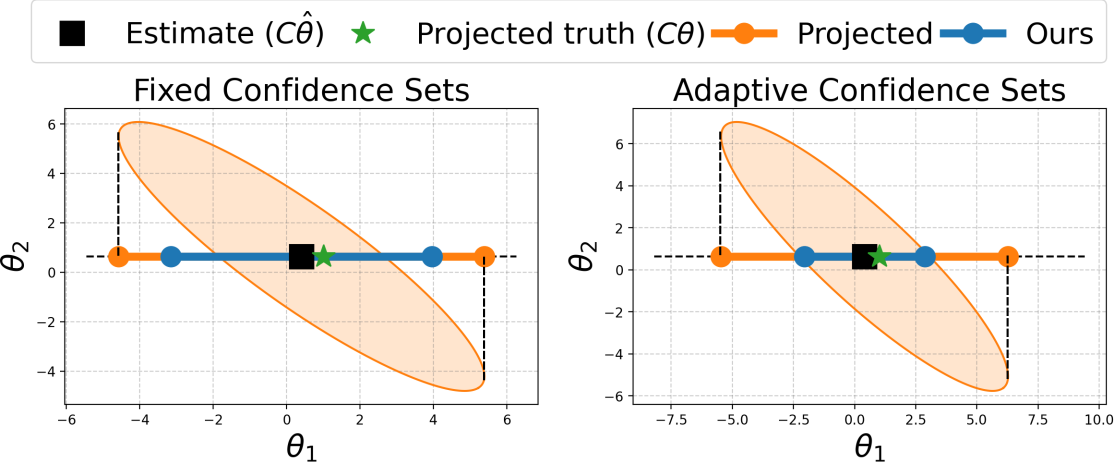}
\caption{A two dim. example. Left: fixed design with our and projected confidence sets from two dimensions. Right: Adaptive confidence sets compared with projected ones due to \citet{Abbasi-Yadkori2012}. In this example $\bC = (1,0)$ and $\theta = 0$.}
\label{fig:ellipse-main}
\end{wrapfigure}

\section{Fixed Designs and their Confidence Sets}\label{sec:bias}
In RKHS spaces, especially infinite dimensional ones, the \emph{estimability} without bias is too restrictive. Given a finite evaluation budget $T$, we can only construct a discrete design $\bX:\mH_\kappa \rightarrow \mR^T$, and there are many practical examples, where, given a finite query budget, $\bC\theta$ cannot be learned to arbitrary precision, most prominently gradients and integrals among many others.

\paragraph{Estimation with Bias and Interpolation Estimator}
Our goal is to establish a condition on the design space $\bX$ such that the estimation with the interpolation estimator is possible up to a certain bias $\epsilon$ measured under the Euclidean norm. We measure the bias scaled by the magnitude of the Frobenius norm of the estimator and call this the relative $\nu$-bias. This condition  will allow us to balance the error due to the bias and noise.

\begin{definition}[Relative $\nu$-bias]\label{def:approx} Let $\bC: \mH_\kappa \rightarrow \mR^k$. The estimator $\bL_\dagger$ on the design space $\bX$ is said to have {\em relative $\nu$-bias} if
	\begin{equation}\label{eq:condition}
		\norm{(\bC - \bL_\dagger \bX)\mV_0^{-1/2}}_{F}^2 \leq \nu^2 \norm{\bL_\dagger}_F^2.
	\end{equation}
\end{definition}
\vspace{-0.2cm}
The $\norm{\cdot}_F$ corresponds to the Frobenius norm of the maps $\mH_\kappa \rightarrow \mH_\kappa$. Due to the cyclic property of the trace, we can take the adjoint of the operator and calculate the quantity by taking a trace of $p \times p$ matrix instead. If $\nu = 0$ and $\dim(\mH_\kappa) = m$, as we show in Lemma \ref{lemma:equivalence-nu-zero} in the Appendix, this is equivalent to the classical \emph{estimability} condition due to \citet{Pukelsheim2006}. The left hand side corresponds to the \emph{classical bias} $\operatorname{bias}(\bL_\dagger y) = \norm{\mE[\bL_\dagger y]-\bC\theta}_2$ of an estimator $\bL_\dagger$ under the Frobenius norm. This is exactly the quantity that $\bL_\dagger$ minimizes. The \emph{classical bias} cannot be improved by repeated measurements or allocations thereof. Using a relation which we show formally in Proposition \ref{prop:equivalence} in Appendix \ref{app:proofs}, we can bound the $\operatorname{bias}(\bL_\dagger y) = \norm{\mE[\bL_\dagger y-\bC\theta]}_2  \leq \lambda_{\min}(\bW_\dagger)^{-1/2}{\nu}/{\sqrt{\lambda}}$.


To check whether the condition \eqref{eq:condition} is satisfied given a design space $\bX$, one needs to evaluate the trace. Finding the value of $p\times p$ matrix, however, depends strongly on the form of the operator $\bC$, and a general recipe cannot be provided. Due to Riesz's representer theorem, and fact that the range of $\bC$ is finite-dimensional, this is always possible. For example, for an integral operator $\int \Phi(x)^\top\cdot q(x)dx$, using shorthand $v = \bK^{-1}y$, it holds that $\bL_\dagger y = \sum_{i=1}^n\int \kappa(x,x_i)v_i dx$ where we used $n$ points. Notice that we used only evaluations of kernel $\kappa$ to do this calculation. Also notice that calculation of \eqref{eq:condition} reduces to the calculation of the maximum mean discrepancy \citep{Gretton2005} between the empirical distribution (defined on $\bX$), and $q(x)$ for this example. The general recipe is to (locally) project the Hilbert space on a truncated finite-dimensional basis $\Phi_m(x)$ with size $m$ and, in that case, the check involves a solution to a linear system, but this is not always necessary.

\subsection{Confidence sets}
To construct confidence sets for both estimators considered, we split them into two categories: \emph{fixed designs}, where the evaluation queries do not depend on the observed values $y$, and  {\em adaptive designs}, where the queries $\Phi(x_i)$ {\em may depend} on prior evaluations $y_{i-1}, \dots y_1$. Proofs are in Appendix~\ref{app:proofs}.

\begin{theorem}[Interpolation -- Fixed Design] \label{thm:fixed_design_pseudo}
	Under the regression model in Eq.~\eqref{eq:model} with $T$ data point evaluations, let $\hat{\theta}$ be the estimate as in \eqref{eq:interpolator}. Let $\bX$  satisfy the bias from Def.~\ref{def:approx} with $\bW_\dagger = (\bC  \mV_0^{-1/2}\bV^\dagger \mV_0^{-1/2} \bC^\top)^{-1}$. Then,
	\begin{equation}
		\pP\left(  \norm{ \bC (\hat{\theta} - \theta) }_{\bW_\dagger} \geq \sigma \sqrt{\xi(\delta)} + \frac{\nu}{\sqrt{\lambda}}   \right) \leq \delta,
	\end{equation}
	where $\bV^\dagger =
	\mV_0^{-1/2}\bX^\top(\bX\mV_0^{-1} \bX^\top)^{-2}\bX\mV_0^{-1/2}$, and $\xi(\delta) = p + 2\left( \sqrt{p\log(\frac{1}{\delta})} +\log\left(\frac{1}{\delta}\right)\right)$.
\end{theorem}
Notice that in order to balance the source of the error due to bias and variance with high probability, we need to match $\sigma \sqrt{\xi(\delta)} \approx \frac{\nu}{\sqrt{\lambda}}$. Repeating the queries in $\bX$ $T$ times reduces $\sigma$ by $1/\sqrt{T}$ but leaves the bias $\nu$, as well as $\bW_{\dagger}$, unchanged. This is because with the interpolation estimator the noisy repeated queries are averaged, which can be interpreted as a reduction in variance. Hence, by balancing $\nu/\sqrt{\lambda}$ with $\sigma \sqrt{\xi(\delta)}/\sqrt{T}$, we balance the bias and variance such that they are of the same magnitude. It does not make sense to repeat measurements more times if the bias dominates the error of estimation. A detailed example of estimating the gradient with fixed bias is given in Sec.~\ref{sec:derivatives}.

We derive confidence sets for the regularized estimator of Eq. \eqref{eq:ridge}, albeit without the relative bias.
\begin{proposition}[Regularized estimate -- Fixed Design] \label{thm:fixed_design_ridge_general}
	Under the model in Eq.~\eqref{eq:model} with $T$ data point evaluations, let $\hat{\theta}_\lambda$ be the regularized estimate as in \eqref{eq:ridge}. Then	$\pP\left(  \norm{ \bC (\hat{\theta}_{\lambda} - \theta) }_{\bW_{\lambda}} \geq  \sqrt{\xi(\delta)} +1\right) \leq \delta,$ where $\xi(\delta) = p + 2\left( \sqrt{p\log(\frac{1}{\delta})} +\log\left(\frac{1}{\delta}\right)\right)$.
\end{proposition}
\looseness -1 Notice that, since the regularized estimator is designed to balance the bias and variance automatically, we do not need to specifically control the bias, which is contained within the information matrix $\bW_{\lambda}$. Despite this elegant property, the regularized estimator involves a more challenging analysis. The main motivation to study the interpolation estimator is to understand the $l_2$ error as we show in Section \ref{sec:complexity}.



\vspace{-0.05cm}
\section{Adaptive Design and Confidence Sets} \looseness -1
To provide confidence sets for adaptively collected data, we need to project the data in $\bX$ onto $\bC$, where we will denote the projection by $\bZ$ further on. With the data projected, we can reason about the reduction of the uncertainty of $\bC\theta$ for each point separately, since to each $\Phi(x_i)$ we can associate a unique $z_i$ in $\mR^p$.

\begin{definition}[Projected data]\label{def:low-rank}
	\looseness -1 Let $z(x)\in \mR^p$, \emph{projected data}, be a vector field  s.t. $ \Phi(x)\mV_0^{-1/2} = z(x) \bC \mV_0^{-1/2} + j(x)$, where $x \in \mS \subseteq \mX$ and $j(x) \in \mH_\kappa$, $|\mS|=n$, such that $\bC \mV_0^{-1/2}j(x) = 0$. 
\end{definition}

Classically, the adaptive confidence sets are understood only for $\bC = \bI$, i.e., the identity \citep{Abbasi-Yadkori2011}. In fact, we can always derive confidence sets for $\bC \theta$ from confidence sets for $\theta$, as they only project the ellipsoid to a smaller dimensional space. However, their resulting size may be unnecessarily large, as the confidence parameter scales as $\mO(\dim(\mH_k))$ in general (see Figure \ref{fig:ellipse-main} for a visual example). 

The martingale analysis of \citet{Abbasi-Yadkori2011} and \cite{delaPena2009} specifically assumes that information matrix $\bV_t$, (where $\bC = \bI$) can be additively decomposed to information matrices due to a single evaluation $\bV_t = \sum_{i=1}^t\Phi(x_i)\Phi(x_i)^\top$ each at a different time. With the matrix $\bW_{\lambda,t}$, this additive decomposition is not always possible. Therefore, to utilize the martingale analysis, which requires this additive property, we consider a different information matrix, which upper bounds $\bW_{\lambda}$. The information matrix we use, $\bOmega_{\lambda}$, is constructed from the projections $z(x_i)$, which have the necessary additive property. It gives rise to confidence sets, where, under the ellipsoidal norm, their size scales as $\Theta(p)$. The estimation error depends on $\bOmega_\lambda$ still, but under this norm, the confidence parameter and information matrix are decoupled in a similar way as for the fixed design.

\begin{theorem}[Ridge estimate -- Adaptive Design] \label{thm:adapt_design_ridge_general}
	Under the regression model in Eq.~\eqref{eq:model} with $t$ adaptively collected data points, let $\hat{\theta}_{\lambda,t}$ be the regularized estimate as in \eqref{eq:ridge}. Further, assume that $\bZ$ is as in Def.~\ref{def:low-rank} where $\bX_t = \bZ_t \bC + \bJ_t\mV_0^{1/2}$. Then for all $t \geq 0$,
	\begin{equation}\label{eq:adaptive}
		\norm{ \bC (\hat{\theta}_{t} - \theta) }_{\bOmega_{\lambda,t}} \leq  \sqrt{2\log\left(\frac{1}{\delta}\frac{\det(\bOmega_{\lambda,t})^{1/2}}{\det(\lambda\bS^{})^{1/2}}\right)}+1
	\end{equation}
	with probability $1-\delta$, where	$\bOmega_{\lambda,t} = \frac{1}{\sigma^2}\bZ_t^\top \bZ_t + \lambda\bS$ and $\bS = (\bC \mV_0^{-1} \bC^\top)^{-1}$.
\end{theorem}
The matrix $\bZ_t$ can be calculated by solving a least-squares problem (projection), whereupon $\bC \mV_{0}^{-1/2} \bJ_t = 0$ as needed by Definition \ref{def:low-rank}. Notice that, on one hand, the above confidence parameter grows only when $\bZ$ is large, but at the same time, the ellipsoid shrinks only in that case as well. Also, the ellipsoid above is necessarily smaller than the one with the information matrix $\bW_{\lambda}$ as  $ \bW_{\lambda} \preceq \bOmega_{\lambda}$ (Lemma \ref{lemma:ordering-adaptive} in Appendix). This means we can use the same confidence parameter to give a bound for $\norm{\cdot}_{\bW_{\lambda,t}}$. Notice that the estimator is the same as before, i.e., $\hat{\theta}_\lambda$, using $\bX$ to define the regression, only the information matrix changes. We also present a visual comparison in Figure \ref{fig:ellipse-main} on a simple example, showing that our fixed and adaptive sets are tighter than projected non-asymptotic sets. Using the shorthand $\beta_t(\delta)$ to define the confidence parameter in \eqref{eq:adaptive}, we can in fact bound the error as
$\norm{\bC(\theta_{\lambda,t}- \theta)}_2 \leq \lambda_{\min}(\bOmega_{\lambda,t})^{-1/2} \beta_t(\delta)$.
It depends on $m = \dim(\mH_\kappa)$ only via $\bOmega_{\lambda,t}$. The dependence of the confidence parameter can be at most $\Theta(\sqrt{\log(1 + T)p})$, see Lemma \ref{lemma:conf-param-size}  in Appendix \ref{app:proofs}. The value of $\lambda_{\min}(\bOmega_{\lambda,t})^{-1}$ depends on the geometry of the set $\{\Phi(x) | x \in \mX\}$ and the projection operator $\bC$ as we show in Section \ref{sec:dimension-dependence}. Note that the lower bound of \cite[
Ex. 20.2.3]{Szepesvari2019} does not apply here, since it makes a statement only about the information matrix $\bW_\lambda$.

\vspace{-0.1cm}
\section{Convex Relaxations, Geometry and Dimensions} \label{sec:complexity}
Suppose we are given a candidate set of experiments, i.e., a unique subset of evaluations $\mS \subset \mX, |\mS| = n$ and a budget $T$ of total queries. We seek an allocation $\bX$, where the rows of $\bX$ contain potentially repeated evaluations from $\mS$. How many times should we repeat each experiment in order to find $\max_{\bX} f(\bW(\bX))$? To address this, experimental design literature relaxes this discrete optimization problem, and optimizes over fractional allocation $\eta \in \Delta^n$, where the number of repetitions for $\Phi(x_i)$ is recovered by rounding $\lceil\eta_i T\rceil$. With this interpretation, the objectives in Sec.~\ref{sec:designs} can be written as
\vspace{-0.15cm}
\begin{equation}\label{eq:opt}
	\eta^* = \argmax_{\eta \in \Delta^n} \left[ f(\bW_\dagger(\bD(\eta)^{1/2}\bX_{\mS})) = f(	\bC(\bV_0^{-1}\bX_{\mS}^\top \bD(\eta)\bX_{\mS}\bV_0^{-1})^\dagger\bC^\top)\right],
\end{equation}
 where $\bD(\eta)$ is the diagonalization operator that produces a diagonal matrix with vector $\eta$ on the diagonal and $\bX_{\mS}$ contains non-repeated elements in $\mS$. We have stated the problem above for the interpolation estimator and $\dim(\mH_k)<\infty$, but it naturally generalizes to kernelized estimators, albeit in a less concise form (see Appendix \ref{app:algorithms}).
\subsection{Optimizing allocations: experiment design algorithms}\label{sec:algorithms} \looseness -1
 Given a subset $\mS$, the problem \eqref{eq:opt} can be approximately solved using either convex optimization methods or a greedy algorithm. A comprehensive review of methods constructing designs $\eta^*$ and rounding techniques to get $\bX$ is \emph{beyond the scope of this work}, and \emph{not the core issue} addressed in this work. We briefly review two versatile approaches  for completeness (more details in Appendix \ref{app:algorithms}).
\vspace{-0.2cm}
\paragraph{Greedy selection} Firstly, one can greedily maximize the scalarized information matrix with the update rule
$\eta_{t+1} = \frac{t}{t+1}\eta_t + \frac{1}{1+t}\delta_t$,
$ \delta_t = \argmax_{x \in \mX} f\left(\bW_{\lambda}\left(\frac{t}{t+1}\eta_t + \frac{1}{t+1}\delta_x\right)\right) $,
where $f$ refers to the scalarization and $\delta_x$ to the discrete measure corresponding to the feature map $\Phi(x)$. Due to the form of the update rule, $t\eta_t$ is always an integer.
\vspace{-0.1cm}
\subsection{Convex optimization} \looseness=-1
Alternatively, convex optimization can provably solve the problem to optimality. Specifically, we look for an allocation using convex optimization methods $\max_{\eta\in \Delta^n} f(\bW_\circ(\bD(\eta)^{1/2}\bX))$ where $ \circ \in \{\dagger,\lambda\}$. Care needs to be taken when selecting $\mS$, as we discuss  in Appendix \ref{app:algorithms}. The most common algorithms for ODE problems are the Frank-Wolfe algorithm \citep{Todd2016} and mirror descent algorithm \citep{Silvey1978}. Additionally, semi-definite reformulation of the above optimization problems can be given as we shown in Appendix \ref{app:SDP}. Optimal designs found via convex optimization need to be rounded in practice. State-of-art rounding techniques are discussed by \citet{Allen-Zhu2017} and \citet{Camilleri2021} for finite and infinite dimensional spaces, respectively. The exhaustive cover of rounding techniques is out-of-scope of the current work. 

\subsection{Estimation error and its dimension dependence}\label{sec:dimension-dependence}

If we were to bound the squared error of estimation in high probability, the importance of the scalarization $\lambda_{\min}(\bW_\dagger)$ becomes apparent. Using the Cauchy Schwarz inequality, we get
\[	\norm{\bC (\theta - \hat{\theta})}_{2} \leq \lambda_{\min}(\bW_\dagger)^{-1/2} 	\norm{\bC (\theta - \hat{\theta})}_{\bW_\dagger} \leq \sqrt{\frac{\lambda_{\min}(\bW_\dagger(\eta^*))^{-1}}{T}} 	(\sigma \sqrt{\xi(\delta)}  + \nu/\sqrt{\lambda} ),\]
\looseness -1 where the term due to Proposition~\ref{thm:fixed_design_pseudo} scales as $\mO(p)$ when properly balanced. Using the optimal allocation $\eta^*$, the number of repetitive evaluations is equal to $\eta^*T$. Inverting the relation above yields a query complexity of the order $T \approx \mO(\frac{p}{\epsilon^2}\lambda_{\min}(\bW_\dagger(\eta^*)^{-1}))$. The optimal value $\lambda_{\min}(\bW_\dagger(\eta^*))$ represents a problem dependent quantity that measures the \emph{difficulty of estimation} and captures the geometry of the set $\{\Phi(x)| x \in \mS\subset \mX\}$. It cannot be bounded in general, but it has an elegant geometric interpretation. 
In particular, it corresponds to the square inverse of the \emph{diameter of the largest inscribed ball in the convex hull of symmetrized $\{\Phi(x) | x \in \mS \}$ in the range of $\bC$} \citep{Pukelsheim1993}.

\looseness -1  At first glance, calculating this quantity might seem complicated, but with an example it is apparent. For example, if $\Phi(x) = x$ s.t. $\norm{x}_2^2\leq 1$ (unit $l_2$ ball in $\mR^m$) and $\bC = v$, where $v$ is a unit vector ($p=1$), the inverse diameter of largest ball we can inscribe in direction of $v$ inside $\norm{x}_2^2\leq 1$ equates to $1$, which is independent of $d$. This is not surprising, since it represents an ``easy'' design space, where for any direction $v$ one can find an action $x$ aligned with that coincides with the optimal design. This is true even if $\bC$ has more rows. On the other hand, if we assume evaluation in the $l_1$ ball $\norm{x}_1\leq 1$ and $\bC$ is a vector of ones (again $p=1$). Then the diameter of the inscribed ball is proportional to the inverse square height of a simplex, namely, $\frac{1}{m}$, despite $p = 1$. Hence, despite the confidence parameter being $\mO(1)$, the complexity depends primarily on the geometry of this set.

To give a more exotic example, if $\bC=\nabla_x\Phi(x)$, and the design space are points with fixed length steps in all unit derection $\{x|\Phi(x\pm e_ih)\}$, where $h$ is the stepsize and $e_i$ are principal vectors in $\mR^d$, we can show that $\lambda_{min}(\bW_\dagger)^{-1} \leq  dh + \mO(h^{2})$, leaving the overall complexity to learn a gradient to scale with $d$ instead of the dimensionality of the RKHS which can be infinite. All formal proofs and references are provided in Appendix \ref{app:algorithm:width}.

\begin{figure*}
	\begin{subfigure}{0.33\textwidth}
		\includegraphics[width=\textwidth]{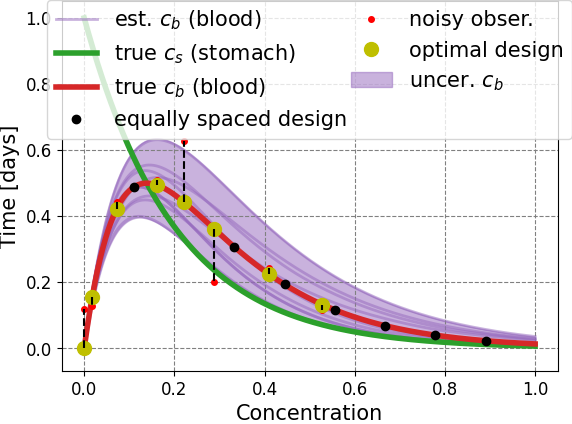}
		\caption{Pharmacokinetics}
		\label{fig:pharma}
	\end{subfigure}
	\begin{subfigure}{0.33\textwidth}
	\includegraphics[width=\textwidth]{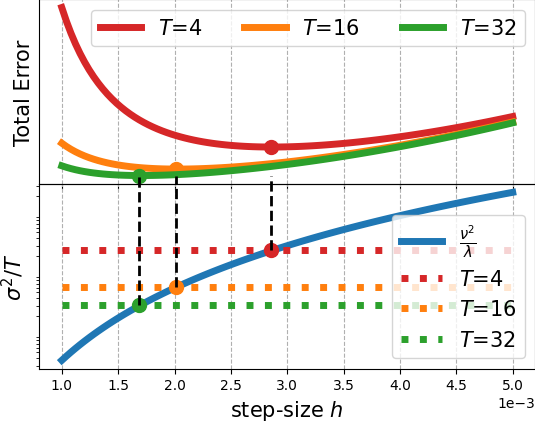}
		\caption{Gradient estimation}
		\label{fig:grads}
	\end{subfigure}
	\begin{subfigure}{0.33\textwidth}
		\includegraphics[width=\textwidth]{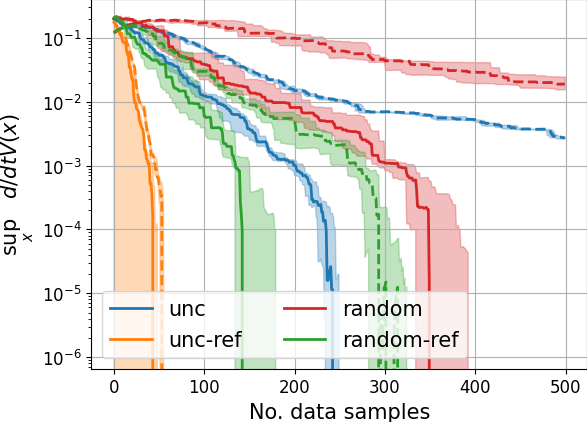}
		\caption{Controller stability}
		\label{fig:control}
	\end{subfigure}

    \vspace{-0.1cm}
	\caption{Experiments: \textbf{a) } \emph{Pharmacokinetics}. We compare the equally spaced design (black) with the optimized design (yellow). The optimized design mimics the classical pharmacokinetics approach of spreading the initial measurements more densely after the initial dose \citep{Gabrielsson1995}. For us, this rule elegantly emerges from first principles. In general, these trajectories follow a decaying pattern as the examples in light purple. The uncertainty in $c_b$  due to the unknown dynamics $\gamma$ is depicted in shaded region. \textbf{ b)} \emph{Gradient estimation}. The upper plot shows the total error of $\nabla f(x)$ we can certify with high probability as a function of the step-size $h$ for a finite difference design. The minima of these errors exactly correspond to  step-sizes derived from the bias-variance trade-off we proposed for the interpolation estimator. With increasing $T$, noise can be reduced more, and hence bias of the design needs to decrease accordingly, hence the decrease in the step size (e.g.,~red$\to$green). \textbf{c)} \emph{Stability:} We report the upper bound on the Lyapunov function in the whole operating domain, as a function of data points. The color coding (explained in the text) represents different data acquisition algorithms while dashed lines correspond to the confidence sets from prior work. Our confidence sets (solid) provide a tighter upper bound as a function of the number of data points, and allows faster termination, which happens when the upper bound is zero. }
	\label{fig:applications}
	\vspace{-0.5cm}
\end{figure*}
\vspace{-0.1cm}
\section{Applications} \label{sec:example} \looseness -1
We now discuss concrete applications that benefit from our contributions. Details of the experiments, and further applications, e.g., in statistical contamination, can be found in Appendices \ref{app:examples} and \ref{app:experiments}. We provide information about optimization and rationale in choosing the $\kappa$, and $\mV_0$ in \ref{app:algorithms} and \ref{app:experiments} respectively. 
\subsection{Linear bandits with finite number of arms}
A special but important theoretical consequence of our adaptive confidence set is an improved regret bound for the upper confidence bound (UCB) algorithm \citep{Auer2002} for linear bandits when the set of queries $\mX$ is finite. The queries are known as actions in the bandit literature \citep{Szepesvari2019}. The UCB algorithm is a procedure which iteratively queries the action according to $x_t = \argmax_{x \in \mX} \max_{\vartheta \in C_{t-1}} \vartheta^\top \Phi(x)$, where $C_t$ is confidence set for $\theta$ in round $t$. Different to this, in our analysis of the UCB algorithm, we use the confidence sets due to Theorem~\ref{thm:adapt_design_ridge_general}, which are confidence sets for the projections $\Phi(x)^\top\theta$ directly. We construct $|\mX|$ confidence sets, one for each of the linear functionals $\bC_x = \Phi(x)$, and denote them by $\operatorname{cf}_t(x)$. The UCB algorithm then just becomes $x_t = \argmax_{x \in \mX}\max_{y \in \operatorname{cf}_t(x)}y$. We now analyze its cumulative regret $R_T:= \sum_{t=1}^T (\Phi(x^*) - \Phi(x_t))^\top \theta$, where $x^* = \argmax_{x\in \mX} \Phi(x)^\top \theta$, and $\theta$ is the true unknown \emph{pay-off vector} of the bandit game. 
\begin{theorem}\label{thm:linear-bandits-main}
	Let $\theta \in \mR^d$ be an unknown pay-off vector, and a set of actions $x \in \mX$ such that $|\mX|<\infty$. Then the cumulative regret of the UCB algorithm is bounded by
	\[	R_T \leq \mO(\sqrt{Td\log (T(|\mX|+1)/\delta)} ) \text{ with } 1-\delta \text{ probability}. \]
\end{theorem}
\vspace{-0.2cm}
\looseness -1 Thus, our adaptive confidence sets for linear functionals yield a regret bound that is 
a factor of $\sqrt{d}$ tighter than the standard bound.
Note that this does not break the known lower bound $\mO(d\sqrt{T})$ for $\mX$ being the unit ball in $\mR^d$ from \citet{Rusmevichientong2010}, as we gain a logarithmic dependence on $|\mX|$. For an $\epsilon$ covering of the extreme points of the unit ball in $\mR^{d}$, one needs $|\mX| \propto (1/\epsilon)^{d}$ points, leading to a regret scaling as $\mO(d\sqrt{T}\log(1/\epsilon)+\epsilon T)$. Choosing $\epsilon = 1/T$ would, for a fixed $T$, recover the lower bound up to logarithmic terms. Using standard \emph{doubling-trick} techniques, one can provide anytime result. However, for different query sets (such as the $l_1$ ball), whose extreme points (i.e., number of vertices) are finite (equal to $d$), these results can vastly improve the regret bounds. There are alternative algorithms based on arm-elimination, which achieve this regret bound, e.g., the algorithm of \citet{Szepesvari2019}, the SupLinRel algorithm of \citet{Auer2002}, and the algorithm of \citet{Valko2014}. The proof of Theorem~\ref{thm:linear-bandits-main}  is deferred to Appendix \ref{app:bandits}, but it is a straightforward application of previous results.

The above technique can be extended to infinite dimensional RKHS, where the role of dimension is played by, $\gamma_T$ called maximum information gain \citep{Srinivas2009}, defined in Appendix \ref{app:bandits}. The remaining challenge is to bound the $\epsilon$-covering of extreme points of the set $\{\Phi(x)| x \in [-1,1]^d\}\subset \mH_k$, where $[-1,1]^d$ is the continuous action set. If all evaluation operators have $\norm{\Phi(x)}_{k} = 1$, for all $x \in [-1,1]$, such as for Mat\'ern kernels \citep{Mutny2018b}, we can use covering numbers of unit balls in these spaces as the upper bound on this number. These spaces are known to be isometrically isomorphic to Sobolev spaces \citep{Wendland2004}, and the bounds on covering numbers of unit balls are known to be of order $(\frac{1}{\epsilon})^{2d/s}$, where $s \in \mathbb{N}$, $s>d/2$, refers to regularity of the Sobolev space \citep{Cucker2002}. This yields a bound on the cumulative regret of the order $\mO(\sqrt{C(d)\gamma_T T}\log T)$, where $C(d)$ can depend on dimension. 

\subsection{Gradient maps} \looseness -1 \label{sec:derivatives}
Gradients of any order are linear operators. We can express the gradient at $x$ as dimension-wise evaluation of the following operator
$ \nabla_x (\Phi(x)^\top\theta) = (\nabla_x \Phi(x))^\top\theta =: \bC \theta. $
Clearly, estimability is nearly always impossible, since any evaluation infinitesimally away from $x$ will be insufficient to eliminate bias. Thus our estimates will invariably be biased for any function with infinite Taylor expansion. Yet, while estimating the gradient from evaluations very close to the original $x$ leads to low bias, it at the same time increases the variance, since the difference in the functional value between the two point evaluations is very small compared to the noise magnitude. Given a finite budget or desired accuracy, the best we can do is to find the best design with optimal bias-variance trade-off given the kernel $\kappa$, budget $T$, and noise variance $\sigma$. We consider a class of parametrized finite difference designs $\{\Phi(x \pm he_i)\}$, where $h$ is the stepsize and $e_i$ are unit vectors. In Fig.~\ref{fig:grads}, we plot the total error with high probability with the budget $T$ as a function of the step-size $h$. Notice that the lowest error occurs exactly when the variance of observations scaled with the confidence parameter is equal to the relative $\nu$ (two lines cross).
\vspace{-0.1cm}
\subsection{Learning linear ODE solutions and their parameters}\label{sec:example:odes}
\looseness -1 A solution to a linear ordinary differential equation (ODE) $u$ satisfies  $\frac{d}{dt} u(t) = \bM u(t) + s(t)$, where $\bM$ is a linear operator and $s(t)$ is the non-homogeneous term. Assume that the solution to the equation $u(t) =\Phi(t)^\top u$ is a member of a Hilbert space $\mH_\kappa$, then $\bT u :=  \left(\frac{d}{dt} - \bM(t) \right) \Phi(t)^\top u = s(t)$ for $ t \in [t_0, t_1].$ Hence, the differential equation becomes a \emph{linear constraint} for estimating  $u$ from samples. In fact, due to differential equations being fully specified by initial conditions, the only unknowns are due to initial conditions. To reveal the linear functional here, one needs to consider the solution to the differential equation, which can be written as $u = \bT^\dagger s(t) + \bC^\top v $, where $v \in \mH_\kappa \setminus \Span(\bT)$ belongs to the null space of $\bT$. In this case, we span it with rows of $\bC$. Consequently, the unknown element $v$ can be found as $\bC (u-\bT^\dagger s)$. Thus, what needs to be estimated from samples, is the linear projection $\bC u$, since $\bC\bT^\dagger s$ is known a priori. In Appendix \ref{app:experiments}, we discuss implementation and calculation of the operator $\bC$ on a discretized domain. Also note that, $\bZ$ itself induces a Hilbert space $\mH_z$ containing the solutions $u$ \citep{Gonzalez2014}. However, as we will see, for robust designs it is sometimes convenient to pick a larger Hilbert space $\mH_\kappa$ s.t. $\mH_z \subset \mH_\kappa$ and then apply the differential equation constraint. 

\vspace{-0.05cm}
\paragraph{Robust Design} \looseness -1 As a specific example, consider an example of a pharmacokinetic model capturing the concentration of a medication in blood and stomach via differential equations: $(d/dt) c_s = -a c_s$ and $(d/dt) c_b = b c_s - d c_b~\text{for} ~ t \in [t_0, t_1]$, where $c_s$ and $c_b$ are the concentration in stomach and blood respectively. The goal of this analysis is to infer $\gamma = (a,b,d)$ from the measurements of the blood concentration levels with fixed, but perhaps noisy, initial conditions \citep{Gabrielsson1995}. To apply the above procedure, we need a fixed differential operator $\bT_\gamma$ to define the operator $\bC_\gamma$. However, $\gamma$ is itself unknown. Instead, we can give a generally plausible set of $\gamma \in \Gamma$, initial conditions of concentration in blood $c_b(0) = 0$, and norm constraints on the initial stomach concentration $(c_s(0)-c_{\text{dose}})^2\leq \lambda^{-1}$ (prior $\mV_0$) in order to apply our framework. We use the squared exponential kernel to embed trajectories and consider robust $A$-optimal design (as in Sec.~\ref{sec:designs}) by maximizing the worst case metric $\inf_{\gamma \in \Gamma}1/\Tr(\bC_\gamma\bV_\lambda^{-1}\bC_\gamma^\top)$ with the regularized estimator. This way, the design is appropriate for any $\gamma$. After estimating the trajectory, we can use the estimated trajectories  (given $\gamma$) to optimize for $\gamma$ via the maximum likelihood. Fig.~\ref{fig:odeint} presents the concentrations $c_b$ and $c_s$ with their estimates and uncertainties due to the unknown $\gamma$. We show the equally spaced  (black) and optimized designs (yellow). The more accurately we can infer the trajectory, the more accurately we can estimate $\gamma$, which we quantitatively show  in Appendix \ref{app:examples} in Fig.~\ref{fig:pharm}, where we decrease the MSE of estimating $\gamma$ by a factor of $10$ in comparison to equally spaced design.

\vspace{-0.1cm}
\subsection{Sequential Design: Certifying Lyapunov Stability}
Consider a non-linear system with $x \in \mR^d$ such that $ \frac{d}{dt}x(t) = f(x(t),t) + u(x(t),t) =
\bA \Phi(x(t),t) + u(x(t),t)$, where the rows of $\bA_i \in \mH_\kappa$ for $i\in [d]$ model the system dynamics and $\Phi(x(t),t)$ are known evaluation functionals of $\mH_\kappa$. We assume that the control laws $u(x(t),t)$ can be written as $u(x(t),t) = \bB \Phi(x(t),t)$. We want to understand whether a given a control law $\bB$ stabilizes the above system. A common approach is to create an estimate $\hat{\bA}$ of $\bA$ from data samples of trajectories, and use the controller $\bB = \hat{\bA} - \bP$, where $\bP$ is often a simple reverting law with a known gain that compensates the imprecise estimation of $\bA$ with $\hat{\bA}$, see below for an example. With this choice of $u$, the system can be written as
$\frac{d}{dt}x(t) = (\bA - \hat{\bA} -\bP)\Phi(x(t),t) = (\tilde{\bA}-\bP)\Phi(x(t),t),$
where the $\tilde{\bA} = \bA - \hat{\bA}$ is the residual error in estimating $\bA$. We can certify the stability of the resulting system using a \emph{known} quadratic Lyapunov function as commonly done in control theory, in this case, using $V(x,t) = (x(t)-x_\text{ref}(t))^\top \bSigma (x(t)-x_\text{ref}(t))$. Classical stability theory dictates that if the total time derivative of $V$, 
\begin{equation}\label{eq:lyapunov-function}
	dV/dt= (x(t)-x_{\text{ref}}(t))^\top \bSigma (\tilde{\bA}-\bP) \phi(x(t),t) +{\partial V}/{\partial t},
\end{equation}
is negative for all $x(t) \in O$ then we can guarantee stability in the operating region of $x \in O$ \citep{Khalil2002}. The above condition defines a linear operator $\bC_x = (x-x_{\text{ref}})^\top \bSigma\cdot\phi(x)$ operating on the unknown $\tilde{\bA}$ for each $x \in O$. The linearity is best seen with vectorization as $\theta = \operatorname{vec}(\bA)$, using the shorthand $z(t) = x(t) - x_{\text{ref}}(t)$,
$ z^\top \bSigma \bA \phi(x) = \operatorname{vec}(\bSigma z\phi(x)^\top)^\top \operatorname{vec}(\bA) = \bC_x \theta.  $
The operator is parametrized by $x$, $\bC_x$ for each $x \in O$. Even for continuous domains $x \in O$, $\bC_x$ usually has low-rank structure, which can be calculated and depends on the size $O$ and the size $\mH_\kappa$. If the rank of the operator is small (e.g., the operating space is small) then we can certify negativity of Eq.~\eqref{eq:lyapunov-function} faster than learning the whole $\bA$. In other words, we reduce uncertainty only where we need to as in the seminal work of \citet{Berkenkamp2016}. We can sequentially query data points from $x$ and check whether $\bC_{x}\theta \leq 0$ for all $x$. 

\looseness-1 Consider a two dimensional nonlinear system from \citet{Lederer2020},
\[ \frac{dx}{dt} = x + \frac{1}{1+\exp(-2x_1)} \begin{pmatrix}1 \\ -1 \end{pmatrix} + 0.5  \begin{pmatrix}\sin(\pi x_2) \\ \cos(\pi x_1) \end{pmatrix} + u. \]
The reverting controller is $\bP\Phi(x(t),t) = -K (x - x_{\text{ref}}(t))+(d/dt)x_{\text{ref}}$, where the reference trajectory corresponds to a circle $x_{\text{ref}}(t) = ( \sin(t), \cos(t))$. The Lyapunov function is $V = (x(t) - x_{\text{ref}}(t))^\top(x(t) - x_{\text{ref}}(t))$. We assume that we can set the system to an initial condition $x(0)$ and observe a noisy observations of the state $y(0+\Delta)$ at rapid sampling times $\Delta$. From these, we create a derivative oracle, $(d/dt) x(t) \approx (x(t+\Delta) - x(t))/\Delta$, a common approach in nonlinear data-driven control \citep{Umlauft2018}. We use this example to showcase our adaptive confidence sets. We follow an adaptive stopping rule, where we query new data points if negativity cannot be certified. Due to this adaptive stopping rule, the data is adaptively collected. 

We compare our confidence sets (solid) with the classical confidence sets of \citet{Abbasi-Yadkori2011} (dashed) and report upper bounds on the supremum over $x \in O$ of Eq.~\eqref{eq:lyapunov-function} in Fig.~\ref{fig:control}. The operating region is a ``tube'' around the circular reference trajectory $x_{\text{ref}}(t)$. We see that our confidence sets shrink much faster than the classical confidence sets, since we can eliminate redundant information by projecting onto $\bC$. Fig.~\ref{fig:control} compares random sampling of datapoints from the whole domain (random) and the operating region (random-ref) and sampling according to uncertainty in the dynamics in the whole domain (unc) and within the tube around the reference trajectory (unc-ref). As expected, focused exploration methods work much better. However, and more importantly, the tightness of our confidence sets enable much quicker stability certification (termination). Note that the classical confidence sets may even {\em grow} in the cases where redundant information for estimation of $\bC$ is inserted into the estimation, e.g., with random sampling.
\vspace{-0.1cm}
\section{CONCLUSION}
\looseness -1 We considered the problem of learning a linear function of an element in a reproducing kernel Hilbert space. We addressed the challenging case where linear estimators incur a non-negligible bias and provided confidence sets for the two most commonly used linear estimators. We demonstrated the generality of our approach and the tightness of our confidence sets on several challenging applications. We believe our results lay important foundations for principled and efficient experiment design in complex real-world settings.

\newpage
\section*{Acknowledgement}
This project has received funding Swiss National Science Foundation through NFP75 and this publication was created as part of NCCR Catalysis (grant number 180544), a National Centre of Competence in Research funded by the Swiss National Science Foundation.

\bibliography{master.bib}
\bibliographystyle{apalike}

\newpage

\appendix
\onecolumn

\begin{center}
{\bf{\Large{Supplementary Material: \\ Experimental Design for Linear Functionals in Reproducing Kernel Hilbert Spaces}}}

\end{center}
\hrule 

\section{Estimability results} \label{app:estimability}

In the following section, we either describe proof of for implication or equivalences of certain conditions studied in this work. In \ref{app:equivalence}, we show consequence of Def. \ref{def:approx} which is used in the proofs of confidence sets. In the subsection following it, we establish relationship to classical ODE as in \citep{Pukelsheim2006} showing that our bias condition generalizes notion stemming from there. 
\subsection{Equivalence of bias conditions}\label{app:equivalence}
\begin{proposition}\label{prop:equivalence}
Let $\bL_{\dagger}$ be the interpolation estimator and $\bW_{\dagger}$ it associated information matrix then, then
\begin{equation}\label{eq:A1}
	\norm{(\bC - \bL_\dagger\bX)\mV_0^{-1/2}}_F^2 \leq  \nu^2 \norm{\bL_{\dagger}}_F^2 \implies	\norm{\bW_{\dagger}^{1/2}(\bC - \bL_\dagger\bX)\mV_0^{-1/2}}_2^2 \leq  \nu^2 
\end{equation}
\end{proposition}
\begin{proof}
	Note that $\bV^\dagger  = \mV_0^{-1/2} \bX^\top \bK^{-2} \bX \mV_0^{-1/2} $, where $\bK = \bX \mV_{0}^{-1} \bX^\top$.

	\begin{eqnarray*}
			& & \text{(LHS)} \\
		&=& \mV_0^{-1/2} (\bC-\bL_\dagger\bX)^\top \bW_\dagger (\bC-\bL_\dagger\bX)\mV_0^{-1/2}\\
		 	&=& \mV_0^{-1/2} (\mI-\mV_0^{-1}\bX^\top\bK^{-1}\bX)^\top \bC^\top (\bC \mV_{0}^{-1/2}\bV^\dagger \mV_{0}^{-1/2} \bC^\top)^{-1} \bC (\mI-  \mV_0^{-1}\bX^\top\bK^{-1}\bX)\mV_0^{-1/2} \\
		 	&=&   \mV_0^{-1/2}(\mI-\bX\bK^{-1}\bX^\top \mV_0^{-1})\bC^\top (\bC  \mV_{0}^{-1/2} \bV^\dagger\bC^\top)^{-1} \mV_{0}^{-1/2} \bC (\mI-\mV_0^{-1}\bX^\top\bK^{-1}\bX)\mV_0^{-1/2}\\
		 	&=&  (\mI-\mV_0^{-1/2}\bX\bK^{-1}\bX^\top \mV_0^{-1/2})\mV_0^{-1/2}\bC^\top (\bC\mV_0^{-1/2}\bV^\dagger\mV_0^{-1/2}\bC^\top)^{-1} \bC \mV_0^{-1/2}(\mI-\mV_0^{-1/2}\bX^\top\bK^{-1}\bX\mV_0^{-1/2})\\
		 	&=&   (\mI-\mP)\mV_0^{-1/2}\bC^\top (\bC \mV_{0}^{-1/2} \bV^\dagger \mV_{0}^{-1/2} \bC^\top)^{-1} \bC \mV_0^{-1/2}(\mI-\mP)\\
	\end{eqnarray*}
		Now let $\tilde{\bX} = \bX\mV_0^{-1/2}$ and $\tilde{\bC} = \bC\mV_{0}^{-1/2}$. Also notice that, $(\tilde{\bX}^\top \tilde{\bX})^\dagger = \mV_0^{-1/2} \bX^\top \bK^{-2} \bX \mV_0^{-1/2} = \bV^\dagger $
		
		We can apply Theorem \ref{thm:main}, to get 
		\[\tilde{\bC}^\top(\tilde{\bC} (\tilde{\bX}^\top \tilde{\bX})^\dagger \tilde{\bC}^\top  )^{-1}\tilde{\bC} = \mV_{0}^{-1/2}\bC^\top(\bC\mV_{0}^{-1/2} (\tilde{\bX}^\top \tilde{\bX})^\dagger \mV_{0}^{-1/2} \bC^\top  )^{-1}\bC\mV_{0}^{-1/2} \preceq c \tilde{\bX}^\top \tilde{\bX} + \mI \nu^2  \]
		Using the above result, we show that 
		\begin{eqnarray*}
			\text{(LHS)} & \preceq & (\mI-\mP) (\mV_{0}^{-1/2}\bX^\top \bX \mV_{0}^{-1/2} + \mI\nu^2) (\mI-\mP)\\
			& = & \nu^2(\mI-\mP)^2 \preceq  \mI \nu^2
		\end{eqnarray*}
		where we have used the fact that $\mI-\mP$ is projection matrix with orthogonal span to $\mV_{0}^{-1/2}\bX^\top \bX \mV_{0}^{-1/2}$.

%
%
%
%
		
%
	
\end{proof}

\begin{proposition}\label{prop:bias-cannot-be-improved}
	Bias $\norm{(\bC - \bL_\dagger\bX)\mV_0^{-1/2}}$ depends only on its support not the allocations. 
\end{proposition}
\begin{proof} \hfill \\
	
	Let $\bL_{\eta}$ be the estimator with design $\eta$, where 
	\begin{eqnarray}
		\bL_\eta & = & \bC \mV_0^{-1} \bX^\top \bD(\eta)^{1/2} (\bD(\eta)^{1/2}\bX \mV_0^{-1} \bX^\top \bD(\eta)^{1/2})^{-1} \\
		& =&  \bC \mV_0^{-1} \bX (\bX \mV_0^{-1} \bX^\top )^{-1}\bD(\eta)^{-1/2} = \bL_{\dagger}\bD(\eta)^{-1/2}	
	\end{eqnarray}
	But at the same time $\bD(\eta)^{1/2}\bX$, hence these two cancel each other. 
	
\end{proof}

\subsection{Basic equivalences and relations }
To relate the estimability to terms used in the ED literature, we state the definition of the \emph{feasibility cone} as in \citet{Pukelsheim2006}. We show that our condition in Def. \ref{def:approx} and Pukelsheims and estimability are equivalent under classical assumptions. We use a different formulation of the relative-bias condition due to Proposition \ref{prop:equivalence}. 

\begin{definition}[\citep{Pukelsheim2006}] \label{def:estimability} Let $\bC: \mH_\kappa \rightarrow \mR^k$, denote $ \mA(\bC) = \{  \bA \in \mP(\mH_\kappa) ~ : \text{range}(\bC^\top) \subseteq \text{range}(\bA)  \} $, where $\mP(\mH_\kappa)$ is the space of positive-definite operators on $\mH_\kappa$. We call $\mA(\bC)$ the feasiblity cone of $\bK$. 
\end{definition}
This definition is sometimes used as restatement of the estimability property. 
\begin{lemma}[Equivalence in $\nu = 0$]\label{lemma:equivalence-nu-zero}
	Let $\bC \in \mR^{k \times d}$ full rank $k$. Let $\bX \in \mR^{n \times d}$, then if $\bM = \bX^\top \bX$, the following are equivalent, 
	\begin{enumerate}
		\item \textbf{Estimability:} There exists $\bL \in \mR^{k \times n}$ s.t.	 $\bC = \bL \bX$.
		\item \textbf{Feasibility cone:} $\bM \in \mA(\bC)$ 
	\end{enumerate}
\end{lemma}
\begin{proof}\hfil
	\begin{itemize}
		\item $(2) \implies (1)$: There exists $\bB$ s.t. $\bC = \bB \bM $, and $\bC = \bB \bX^\top \bX$, hence we can define $\bL = \bB \bX$. 
		\item $(1) \implies (2)$: As such $\bC = \bL \bX$ implies that $\text{range}(\bC) \subseteq \text{range} (\bX^\top) = \text{range} (\bX^\top \bX)$.
	\end{itemize}
\end{proof}

\begin{definition}[Projected data]\label{def:low-rank2}
	\looseness -1 Let $z(x)\in \mR^p$ be a vector field  s.t. $ \Phi(x)\mV_0^{-1/2} = z(x) \bC \mV_0^{-1/2} + j(x)$, where $x \in \mS \subseteq \mX$ and $j(x) \in \mH_\kappa$, $|\mS|=n$, such that $\bC \mV_0^{-1/2}j(x) = 0$. We call this vector field \emph{projected data}. If in addition if $\{z(x):x\in \mS\}$ spans $\mR^p$ it is said to be \emph{approximately low-rank}.
\end{definition}

\begin{lemma}\label{lemma:implication} The assumption in Definition \ref{def:low-rank2} implies the assumption in Definition \ref{def:approx} with $\nu = \norm{\bJ}_{k} \frac{\norm{\bZ^\dagger}_F}{\norm{\bL_\dagger}_F}$.
\end{lemma}
\begin{proof}
	Since $\bZ$ spans whole $\mR^k$, there exists a unique left pseudo-inverse. $\bZ^\dagger \bX \mV_0^{-1/2} = \bC\mV_0^{-1/2} + \bZ^\dagger \bJ$. 
	\[  (\bC - \bZ^\dagger \bX)\mV_{0}^{-1/2} = \bZ^\dagger \bJ \]	
	Now taking the Frobenius norm, 
	\[
	\norm{(\bC - \bZ^\dagger \bX)\mV_{0}^{-1/2}}_{F} \leq \norm{\bZ^\dagger \bJ }_F = \norm{\bJ}_{F}\norm{\bZ^\dagger}_F
	\]
	So, 
	\[
	\frac{\norm{(\bC - \bZ^\dagger \bX)\mV_{0}^{-1/2}}_{F}}{\norm{\bZ^\dagger}_F} \leq \norm{\bJ}_{F}
	\]
	
	Now, since pseudo-inverse minimizes the LHS, we know that, 
	\[\norm{(\bC - \bL_\dagger \bX)\mV_{0}^{-1/2}}_{F}  \leq	\norm{(\bC - \bZ^\dagger \bX)\mV_{0}^{-1/2}}_{F} \leq \norm{\bJ}_{k} \frac{\norm{\bZ^\dagger}_F}{\norm{\bL_\dagger}_F}{\norm{\bL_\dagger}_F} \]
\end{proof}
\section{Confidence Sets: Proofs}\label{app:proofs}
This section includes proofs for the concentration results presented in the main text. In Section, \ref{app:relation-prior} we restate and prove results from \citet{Mutny2020} in the current notation for easier comparison. 
\subsection{Fixed Design with Interpolator}
The following Theorem tries to qualify whether a design contains sufficient information to reduce confidence on a specific subspace $\bC$ to the desired error.

\begin{theorem}[Interpolation (Fixed Design)] \label{thm:fixed_design_pseudo2}
	Under regression model in Eq.~\eqref{eq:model} with $T$ data point evaluations, let $\hat{\theta}$ be the estimate as in \eqref{eq:interpolator}. Further let $\bX$ be s.t. satisfy the approximate estimability condition of Def.~\ref{def:approx} with $T\geq k$. Then,
	\begin{equation}
		\pP\left(  \norm{ \bC  (\hat{\theta} - \theta) }_{\bW} \geq \sigma \sqrt{\xi(\delta)} + \frac{\nu}{\sqrt{\lambda}}   \right) \leq \delta,
	\end{equation}
	where $\bW = (\bC  \mV_0^{-1/2} \bV^\dagger  \mV_0^{-1/2} \bC^\top)^{-1}$, $\bV^\dagger = \mV_0^{-1/2} \bX^\top \bK^{-2} \bX \mV_0^{-1/2} $, and $\xi(\delta) = p + 2\left( \sqrt{p\log(\frac{1}{\delta})} +\log\left(\frac{1}{\delta}\right)\right)$.
\end{theorem}
\begin{proof}
	The pseudo inverse estimate of $\bC\theta$ is $\bC\hat{\theta} = \bC\mV_0^{-1}\bX^\top (\bX\mV_0^{-1}\bX^\top)^{-1}y $, where $\bX \in \mR^{T \times m}$.
	\begin{eqnarray*}
		\norm{ \bC (\hat{\theta} - \theta) }_{\bW}^2  & = & \norm{ \bC  (\mV_0^{-1}  \bX^\top (\bX \mV_0^{-1} \bX^\top)^{-1} (\bX \theta + \epsilon)    - \theta) }_{\bW} ^2  \\ 
		& =&  \norm{ \bC(\mV_0^{-1}  \bX^\top (\bX \mV_0^{-1} \bX^\top)^{-1} \bX - \mI )\theta + \bC \mV_0^{-1}( \bX^\top (\bX \mV_0^{-1} \bX^\top)^{-1} \epsilon)}_{\bW}^2  \\
		& \leq & \norm{ \bC(\mV_0^{-1}  \bX^\top (\bX \mV_0^{-1} \bX^\top)^{-1} \bX - \mI )\mV_0^{-1/2}\mV_0^{1/2}\theta}_{\bW}^2 \\ & & +     \norm{\bC\mV_0^{-1}  \bX^\top (\bX\mV_0^{-1} \bX^\top)^{-1}\epsilon}_{\bW}^2  \\
		& \leq & \norm{ \bW^{1/2}\bC(\mV_0^{-1}  \bX^\top (\bX \mV_0^{-1} \bX^\top)^{-1} \bX - \mI )\mV_0^{-1/2}}_2^2\norm{\theta}_{\mV_0} \\ & & + \norm{\bC\mV_0^{-1}  \bX^\top (\bX\mV_0^{-1}\bX^\top)^{-1}\epsilon}_{\bW}^2 \\
		& = & \lambda^{-1}\nu^2 + \norm{\bC\mV_0^{-1}  \bX^\top (\bX\mV_0^{-1}\bX^\top)^{-1}\epsilon}_{\bW}^2 \\
	\end{eqnarray*}

	where the second to last line we used the relative-bias assumption, according to the Proposition \ref{prop:equivalence}, where we use the Def.~\ref{def:approx} and $T\geq k$. The operator $\mI$ is identify operator on $\mH_k$.


	
	If $\epsilon_G$ were Gaussially distributed then, $\bC  \mV_0^{-1}\bX^\top (\bX\mV_0^{-1}\bX^\top)^{-1}\epsilon_G$ is distributed as $\mN(0, \bC \bV^\dagger \bC^\top)$. 
	Likewise, $q_G = \bW^{1/2}\bC \mV_0^{-1/2} \bX^\top (\bX\mV_0^{-1}\bX^\top)^{-1}\epsilon_G$ is distributed as $\mN(0, \bI_k)$. Since $\epsilon$ is sub-Gaussian $q$ needs to have tails of distribution which are below the tails of $k$-dimensional gaussian. In other words $P(|q_i|\leq t ) \leq P(|q_{G_i}| \leq t)$. As such, $P(q_i^2 \leq t^2) \leq P(|q_{G_i}|^2 \leq t)$, where $q_{G_i}^2$ is chi-squared distributed. 
	Using concentration for chi-squared random variables, $\pP\big(\norm{z}_2^2 - \sigma^2 p\geq\sigma^2( 2px +  2\sqrt{px})\big) \leq \exp(-x)$ as in \cite{Laurent2000}. Rearranging the expression leads to the result with $\xi(\delta) = (p + 2\log(\frac{1}{\delta})+ 2 \sqrt{p\log(\frac{1}{\delta})})$, multiplied by $\sigma^2$.

	The last final inequality in the statement of the probability follows from taking a square root and triangle inequality, which finishes the proof.
\end{proof}


\subsection{Fixed Design with Regularized Estimator}

\begin{proposition}[Fixed Design Ridge Regression]
	Under regression model in Eq.~\eqref{eq:model} with $T$ data point evaluations, let $\hat{\theta}_\lambda$ be the regularized estimate as in \eqref{eq:ridge}. Then,
	\begin{equation*}
		\pP\left(  \norm{ \bC (\hat{\theta}_{\lambda} - \theta) }_{\bW_{\lambda}} \geq \sqrt{\beta(\delta)} = \sqrt{\xi(\delta)} + 
	  1 \right) \leq \delta,
	\end{equation*}
	where $\bW_{\lambda} = \sigma^{-2}(\bC \mV_{\lambda}^{-1} \bC^\top)^{-1} $, $\mV_\lambda^{-1} = (\bX^\top \bX + \lambda \sigma^2 \mV_0)^{-1}$. In addition, $\xi(\delta) = p + 2\left( \sqrt{p\log(\frac{1}{\delta})} +\log\left(\frac{1}{\delta}\right)\right)$.
	
\end{proposition}

\begin{proof}
	Notice that invertibility of $\bW$ is guaranteed by full rank $\bC$ and invertibility of $\bV_{\lambda}$. Using shorthand $\mV = \bX^\top \bX$,
	\begin{eqnarray*}
		\norm{ \bC (\hat{\theta}_\lambda - \theta) }_{\bW_\lambda }^2  & = & \norm{ \bC \mV_0^{-1} \bX^\top (\bX\mV_0^{-1}\bX^\top + \sigma^2\lambda\bI)^{-1} (\bX \theta + \epsilon)    - \theta) }_{\bW_\lambda }^2 \\
		& = & \norm{ \bC \mV_0^{-1/2}  (\mV_0^{-1/2}\bX^\top\bX\mV_0^{-1/2} + \sigma^2\lambda\bI)^{-1} \mV_0^{-1/2}\bX^\top (\bX \theta + \epsilon)    - \theta) }_{\bW_\lambda }^2 \\
		& = & \norm{ \bC\mV_{\lambda}^{-1}\bX^\top (\bX \theta + \epsilon)    - \theta) }_{\bW_\lambda }^2 \\
		& = &  \norm{ \bC ( \mV_\lambda^{-1} \bX^\top \epsilon + \mV_\lambda^{-1}\mV \theta - \theta)}_{\bW_\lambda}^2   \\
		& \leq & \norm{ \bC \mV_\lambda^{-1} \bX^\top\epsilon}_{\bW_\lambda }^2 +  \norm{\bC(\mV_\lambda^{-1}\bX^\top \bX - \mI)\theta}^2_{\bW_\lambda } \\
	\end{eqnarray*}
	
	The second term, 
	\begin{eqnarray}
\  \norm{\bC\mV_\lambda^{-1}(\mV_0 \lambda \sigma^2)\theta)}_{\bW_\lambda}^2 
& 	\stackrel{\eqref{eq:ranking}} \leq & \sigma^2  \norm{\mV_\lambda^{-1}(\mV_0 \lambda )\theta)}_{\mV_\lambda}^2 \\
& = &   \lambda \sigma^2 \theta^\top \mV_0  \mV_\lambda^{-1} (\mV_0 \lambda )\theta) \leq \lambda \theta^\top \mV_0 \theta \leq 1. 
\end{eqnarray}
	
	Let us analyze the first term, 
	\begin{eqnarray}
		\norm{ \bC \mV_\lambda^{-1} \bX^\top\epsilon}_{\bW_\lambda }^2  &=& \epsilon^\top \bX \mV_{\lambda}^{-1}\bC^{\top}  \bW_\lambda \bC \mV_{\lambda}^{-1}\bX^\top \epsilon 
	\end{eqnarray}
	
	The distribution of $\bC\mV_{\lambda}^{-1}\bX^\top \epsilon$ is $\mN(0,\sigma^2\bC \mV_{\lambda}^{-1}\bV \mV_{\lambda}^{-1} \bC^\top)$. Further, $(\sigma\bC \mV_{\lambda}^{-1}\bC^\top)^{-1/2}\bC\mV_{\lambda}^{-1}\bX^\top \epsilon$ is distributed as $z \sim \mN(0,(\bC \mV_{\lambda}^{-1}\bC^\top)^{-1/2} \bC \mV_{\lambda}^{-1}\mV \mV_{\lambda}^{-1} \bC^\top(\bC \mV_{\lambda}^{-1}\bC^\top)^{-1/2} )$.  Let us call the covariance matrix $\bSigma$. It is easy to see that $\bSigma \preceq \bP$, an projection matrix with $k$ unit eigenvalues.The random variable $z$ can be generated as $z = q^\top \bSigma q$, where $q \sim \mN(0, \bI_{n})$. Our goal is to bound,$ P(\norm{z}_2^2 \geq t)$. Using eigenvalue decomposition of $\bSigma = \bQ \bLambda \bQ^\top$, we can show
	
	\[\pP(\norm{z}_2^2 \geq t) = \pP( q^\top \bQ \bLambda \bQ^\top  q^\top \geq t) = \pP( q^\top \bLambda  q^\top \geq t) \leq \pP( q^\top \bP q^\top \geq t) = \pP( \norm{\tilde{q}}_2^2 \geq t)\]
	where $\tilde{q}$ is $p$-dimensional standard normal. Using concentration for chi-squared random variables, $\pP\big(\norm{z}_2^2 - \sigma^2 p\geq\sigma^2( 2px +  2\sqrt{px})\big) \leq \exp(-x)$ as in \cite{Laurent2000}. Rearranging the expression leads to the result with $\xi(\delta) = (p + 2\log(\frac{1}{\delta})+ 2 \sqrt{p\log(\frac{1}{\delta})})$, multiplied by $\sigma^2$.  To deal with sub-Gaussianity we apply the same trick as in the previous Theorem.

\end{proof}

\subsection{Adaptive design and Regularized Estimator}
\begin{theorem}[Adaptive Design Regularized Regression]
	Under the regression model in Eq.~\eqref{eq:model} with $t$ adaptively collected data points, let $\hat{\theta}_{\lambda,t}$ be the regularized estimate as in \eqref{eq:ridge}. Further, assume that $\bZ$ is as in Def.~\ref{def:low-rank} where $\bX_t = \bZ_t \bC  + \bJ_t\mV_0^{1/2}$. Then for all $t \geq 0$,
	\begin{equation}\label{eq:adaptive2}
		 \norm{ \bC (\hat{\theta}_{t} - \theta) }_{\bOmega_{\lambda,t}} \leq  \sqrt{2\log\left(\frac{1}{\delta}\frac{\det(\bOmega_{\lambda,t})^{1/2}}{\det(\lambda\bS)^{1/2}}\right)}+1
	\end{equation}
	with probability $1-\delta$, where	$\bOmega_{\lambda,t} = \frac{1}{\sigma^2}\bZ_t^\top \bZ_t + \lambda\bS$ and $\bS = (\bC \mV_0^{-1} \bC^\top)^{-1}$.
\end{theorem}

\begin{proof}
	
	Notice that the theorem is stated with $\bS$ and $\bS^{-1}$ having inverted roles in the proof, however the result above follows by swapping the two.

	Note that $\bJ\mV_0^{-1/2}\bC^\top = 0$ as well as $\bC\mV_0^{-1/2}\bJ^\top = 0$ due to the assumption. Also, notice that $\bX = \bZ \bC + \bJ \mV_0^{1/2}$. We drop the $t$ subscript for brevity. 
	
	\begin{eqnarray*}
		\norm{ \bC \hat{\theta} - \bC\theta}_{\bOmega_\lambda }^2  & = & \norm{ \bC ( \mV_\lambda^{-1} \bX^\top (\bX \theta + \epsilon)    - \theta) }_{\bOmega_\lambda }^2  =  \norm{ \bC ( \mV_\lambda^{-1} \bX^\top\epsilon + \mV_\lambda^{-1}(\mV - \mV_{\lambda})\theta)}_{\bOmega_\lambda}^2   \\
		& \leq &  \norm{ \bC \mV_\lambda^{-1} \bX^\top\epsilon}_{\bOmega_{\lambda}}^2 + \norm{\bC\mV_\lambda^{-1}(\mV - \mV_{\lambda})\theta)}_{\bOmega_\lambda}^2
	\end{eqnarray*}

	Now we analyze the two terms separately, The first term,
	
	\begin{eqnarray*}
		\norm{ \bC  \mV_\lambda^{-1} \bX^\top\epsilon}_{\bOmega_{\lambda}}^2 & = &  \norm{ \bC \mV_0^{-1} \bX^\top (\bX\mV_0^{-1}\bX^\top + \sigma^2\lambda\bI)^{-1} \epsilon}_{\bOmega_{\lambda}}^2 \\
		& = &  \norm{ \bC \mV_0^{-1} (\bZ\bC+\bJ\mV_0^{1/2})^\top (\bX\mV_0^{-1}\bX^\top + \sigma^2\lambda\bI)^{-1} \epsilon}_{\bOmega_{\lambda}}^2 \\
		& \leq & \norm{ \bC \mV_0^{-1} (\bZ\bC)^\top (\bX\mV_0^{-1}\bX^\top + \sigma^2\lambda\bI)^{-1} \epsilon}_{\bOmega_{\lambda}}^2 \\ & & + \norm{ \underbrace{\bC \mV_0^{-1/2} \bJ^\top}_{=0} (\bX\mV_0^{-1}\bX^\top + \sigma^2\lambda\bI)^{-1} \epsilon}_{\bOmega_{\lambda}}^2 \\
		& = & \norm{ \bC \mV_0^{-1}\bC^\top \bZ^\top ((\bZ \bC +\bJ\mV_0^{1/2})\mV_0^{-1}(\bZ \bC +\bJ\mV_0^{1/2})^\top + \sigma^2\lambda\bI)^{-1} \epsilon}_{\bOmega_{\lambda}}^2 \\
		& = & \norm{ \bC \mV_0^{-1}\bC^\top \bZ^\top (\bZ \bC\mV_0^{-1}\bC^\top \bZ + \underbrace{\bJ\mV_0^{-1/2}\bC}_{=0}\bZ^\top + \bZ \underbrace{\bC\mV_0^{-1/2}\bJ^\top}_{=0} + \bJ\bJ^\top \sigma^2\lambda\bI)^{-1} \epsilon}_{\bOmega_{\lambda}}^2 \\
		& = & \norm{ \bC \mV_0^{-1}\bC^\top \bZ^\top (\bZ \bC\mV_0^{-1}\bC^\top \bZ  + \bJ\bJ^\top \sigma^2\lambda\bI)^{-1} \epsilon}_{\bOmega_{\lambda}}^2 \\
		& \leq & \norm{ \bC \mV_0^{-1}\bC^\top \bZ^\top (\bZ \bC\mV_0^{-1}\bC^\top \bZ  + \sigma^2\lambda\bI)^{-1} \epsilon}_{\bOmega_{\lambda}}^2
	\end{eqnarray*}
	Let us define shorthand $\bS^{-1} = \bC \mV_0^{-1} \bC^\top$ which is $\mR^{k \times k}$ p.s.d. matrix. 
	\begin{eqnarray}
			\norm{ \bC  \mV_\lambda^{-1} \bZ^\top\epsilon}_{\bOmega_{\lambda}}^2 & \leq & \norm{ \bS^{-1/2}\bS^{-1/2} \bZ^\top (\bZ \bS^{-1/2}\bS^{-1/2} \bZ  + \sigma^2\lambda\bI)^{-1} \epsilon}_{\bOmega_{\lambda}}^2	\\
		& = & \norm{ \bS^{-1/2}(\bS^{-1/2}\bZ^\top \bZ\bS^{-1/2}  + \sigma^2\lambda\bI)^{-1} \bS^{-1/2} \bZ^\top \epsilon}_{\bOmega_{\lambda}}^2 \\
		& = & \norm{ (\bZ^\top \bZ  + \sigma^2\lambda\bS)^{-1} \bZ^\top \epsilon}_{\bOmega_{\lambda}}^2 \\
		& = & \norm{ \bZ^\top \epsilon}_{\bOmega_{\lambda}^{-1}}^2
	\end{eqnarray}

	The term above is so called self-normalized noise, which can be handled by techniques of \citet{delaPena2009} popularized by \citet{Abbasi-Yadkori2011}. From now on the proof is generic. Let us define, the noise process $S_t = \sum_{i=1}^{t}z_i \frac{\epsilon_i}{\sigma^2}$ and variance $ \bV_t = \frac{z_i z_i^\top}{\sigma^2} $. Also, let $M_t(x) = \exp(S_t^\top x - \frac{1}{2}x^\top \bV_t x)$ be a process with index $t$. 
	
	Due to sub-gaussianity of $\epsilon_i$, $M_t(x)$ is a super-martingale under filtration that includes all $x_{t-1}, \epsilon_{t-1}, \dots x_1, \epsilon_1$. Now it is worth noting that this all has been introduces since $\exp(\norm{ \bZ^\top \epsilon}_{\bOmega_{\lambda}^{-1}}^2) = \sup_{x} M_t(x)$. This relation allow us to see that if we can upper bound $\sup_{x} M_t(x)$ we get what we want. We proceed by pseudo-maximization. We pick a fixed probability distribution $h(x)$ and define a process $\bar{M}_t =  \int M_t(x) h(x) dx$. Since, $h(x)$ is fixed and normalized $\bar{M}_t$ is also a super-martingale. It turn out that $h(x) \sim \mN(0, \lambda^{-1}\bS^{-1})$ will achieve the desired result.

	\begin{eqnarray*}
			\bar{M}_t & = & \int M_t(x)h(x)dx = \frac{1}{\sqrt{(2\pi)^{k} \det((\lambda )^{-1}\bS^{-1})}}\int_{\mR^k}\exp( x^\top S_t - \frac{1}{2} \norm{x}_{\bV_t} - \frac{\lambda }{2} \norm{x}_{\bS^{}})dx \\
			& = & \left(\frac{\det(\bOmega_{\lambda,t})}{\det(\bS)}\right)^{1/2} \exp(\frac{1}{2}\norm{S_t}^2_{\bOmega_{\lambda,t}})
	\end{eqnarray*}
where we have applied standard rules for Gaussian integrals. Now, we use Ville's martingale inequality to bound,

\begin{eqnarray}
	P\left(\sup \bar{M}_t \geq \frac{1}{\alpha}\right) &\leq &\mE[\bar{M}_t]\alpha = \alpha\\
	P\left(\log(\sup \bar{M}_t) \geq \log\left(\frac{1}{\alpha}\right)\right) &\leq& \alpha \\ 
	P\left(\norm{S_t}_{\bOmega_{\lambda,t}^{-1}}^2  \geq 2\log\left(\frac{1}{\alpha}\right) + 2\log\left(\left(\frac{\det(\bS)}{\det(\bOmega_{\lambda,t})}\right)^{-1/2}\right)\right)&\leq&  \alpha \\
	P(\norm{S_t}_{\bOmega_{\lambda,t}^{-1}}^2  \geq  2\log\left(\frac{1}{\alpha}\det\left(\frac{\det(\bS)}{\det(\bOmega_{\lambda,t})}\right)^{-1/2}\right)&\leq&  \alpha \\
\end{eqnarray} 
which finishes bounding the first term by $2\log\left(\frac{1}{\alpha} \left(\frac{\det(\bOmega_{\lambda,t})}{\det(\bS)}\right)^{1/2}\right)$

Now we turn to the second term, 

\begin{eqnarray}
\norm{\bC\mV_\lambda^{-1}(\mV - \mV_{\lambda})\theta)}_{\bOmega_\lambda}^2 & = & \norm{\bC\mV_\lambda^{-1}(\mV_0 \lambda \sigma^2)\theta)}_{\bOmega_\lambda}^2	\\
& \stackrel{\eqref{eq:ordering-adaptive}} \leq &  \norm{\bC\mV_\lambda^{-1}(\mV_0 \lambda \sigma^2)\theta)}_{\bW_\lambda}^2 \\
& 	\stackrel{\eqref{eq:ranking}} \leq &  \norm{\mV_\lambda^{-1}(\mV_0 \lambda )\theta)}_{\mV_\lambda}^2 \\
& = &   \lambda \theta^\top \mV_0  \mV_\lambda^{-1} (\mV_0 \lambda )\theta) \leq \lambda^2 \theta^\top \mV_0 \theta \leq 1 \\
\end{eqnarray}
This proves the result. 
\end{proof}

\begin{lemma}[Ordering] \label{lemma:ordering-adaptive}
	\begin{equation}\label{eq:ordering-adaptive}
	 \bW_\lambda \preceq \bOmega_{\lambda}
	\end{equation}

\end{lemma}
\begin{proof} Consider series of psd. manipulations,
	\begin{eqnarray*}
		\bC^\top \bW_\lambda \bC & = &    \sigma^{-2} \bC^\top( \bC(\bX^\top \bX  + \sigma^2 \lambda \mV_0 )^{-1}\bC^\top   )^{-1}  \bC \\
&		\stackrel{\eqref{eq:ranking}} \preceq   & \sigma^{-2}\bX^\top \bX  + \lambda\mV_0 \\ &=& \sigma^{-2} (\bC^\top \bZ^\top \bZ \bC + \bC^\top \bZ^\top \bJ \mV_0^{1/2}+ \mV_0^{1/2}\bJ^\top \bZ \bC + \mV_0^{1/2}\bJ^\top \bJ\mV_0^{1/2})\\ & & + \lambda\mV_0 \\
	\bC \mV_0^{-1}\bC^\top \bW_\lambda \bC\mV_0^{-1} \bC^\top & \preceq & \bC  \mV_0^{-1}( \sigma^{-2}\bC^\top \bZ^\top \bZ \bC + \sigma^{-2} \bC^\top \bZ^\top \bJ \mV_0^{1/2} \\ & & +  \sigma^{-2}\mV_0^{1/2}\bJ^\top \bZ \bC +  \sigma^{-2}\mV_0^{1/2}\bJ^\top \bJ\mV_0^{1/2} +  \lambda\mV_0)\mV_0^{-1}\bC^\top \\
	& = & \sigma^{-2}\bS^{-1} \bZ^\top \bZ \bS^{-1} + \lambda \bC\mV_0^{-1} \bC^\top \\
	& = & \sigma^{-2}\bS^{-1} \bZ^\top \bZ \bS^{-1} + \lambda \bS^{-1} \\
\bW_\lambda	& \preceq & \frac{\bZ^\top\bZ}{ \sigma^2 } +\lambda\bS = \bOmega_{\lambda}
	\end{eqnarray*}
	where we use the fact that $\bC\mV_0^{-1/2}\bJ^\top = 0$. 
\end{proof}

\begin{lemma}[confidence parameter size] \label{lemma:conf-param-size} Under assumptions of Thm.~\ref{thm:adapt_design_ridge_general}, where $\norm{\Phi(x)}_k^2 \leq L^2 $, $\mV_0 = \mI$,
\[\sqrt{2\log\left(\frac{1}{\delta}\frac{\det(\bOmega_{\lambda,t})^{1/2}}{\det(\lambda\bS)^{1/2}}\right)} \leq \sqrt{p\log\left( \frac{tL^2}{p\lambda}  + 1 \right) + 2 \log(1/\delta)} = \mO(\sqrt{p \log(t/p)})\]
Further, if $\norm{\Phi(x)}_k \approx (\sqrt{m})$ grows with $m$, e.g. $\Phi(x) = 1_m$ (vector of ones) then 
\[\sqrt{2\log\left(\frac{1}{\delta}\frac{\det(\bOmega_{\lambda,t})^{1/2}}{\det(\lambda\bS)^{1/2}}\right)} = \mO(\sqrt{p\log(tm/p\lambda + 1)}) \]
\end{lemma}
\begin{proof}
First let us determine the absolute bound on all $\norm{z_i}_2^2$. Due to projected data, 

\[ \bS^{-1/2}z_i = (\bC \bC^\top)^{1/2} (\bC \bC^\top)^{-1}\bC \Phi(x_i) \implies \] \[\norm{\bS^{-1/2}z_i}_2^2 \leq \norm{(\bC \bC^\top)^{1/2}(\bC \bC^\top)^{-1}\bC \Phi(x_i)}_2^2 \leq \norm{\bC^\top (\bC \bC^\top)^{-1}\bC}^2_2 L^2 = L^2. \]
where the last step follows from properties of projection.
Notice also that $\frac{1}{\sigma^2}\Tr(\bS^{-1/2}\bZ^\top \bZ\bS^{-1/2}) = \frac{1}{\sigma^2}\sum_{i=1}^{t}\Tr(\bS^{-1/2}z_iz_i^\top\bS^{-1/2}) \leq t\frac{L^2}{\sigma^2}.$
	
Using arithmetic-geometric mean inequality as in \citep{Szepesvari2019} Lemma 19.4, we can show that
\begin{eqnarray*}
 \frac{\det(\bOmega_{\lambda,t})}{\det(\lambda \bS)} & = & \frac{\det(\bS^{-1/2}\bOmega_{\lambda,t}\bS^{-1/2})}{\det(\lambda \bI)} \\
 & \leq &  \frac{(\frac{1}{p} \Tr(\det(\bS^{-1/2}\bOmega_{\lambda,t}\det(\bS^{-1/2}) )^{p}}{\lambda^p} =\frac{(\frac{1}{p} \Tr(\bS^{-1/2}\bZ\bZ^\top\bS^{-1/2} + \lambda \bI) )^{p}}{\lambda^p} \\ 
 &\leq&  \frac{(\frac{1}{p} (t L^2 + \lambda p))^p }{\lambda^p }  \\
 \log\left(   \frac{\det(\bOmega_{\lambda,t})}{\det(\lambda \bS)}  \right) & \leq & p \log\left( \frac{tL^2}{p\lambda}  + 1 \right)
\end{eqnarray*}
The last line of the Lemma follows from noting  that $L^2 = \mO(m)$. 

\end{proof}

\subsection{Adaptive design and Modified Regularized Estimator: Relation to prior work}\label{app:relation-prior}
With the adaptive estimator, we could alternatively directly define an estimator for $\bC\theta$ as $\bC\hat{\vartheta}$ using only the projected values $\bZ$ as done by \citet{Mutny2020}. In this case, however, additional bias growing in time $t$ enters into the confidence parameter as \eqref{eq:adaptive} as $\frac{1}{\lambda}\sum_{i=1}^{t} \norm{j_i}_{\mV_0^{-1}}$, as we show in Theorem \ref{thm:adaptive-bias} below. 

\begin{theorem}[Adaptive Design Regularized Regression - biased]\label{thm:adaptive-bias}
	Let $\bZ_t$ be s.t. $\bX_i = \bZ_i\bC + \bJ_i$, where $\bJ_i$ is minimal as measured by squared norm, then the estimator
	
	\begin{equation}
		\bC\hat{\vartheta}_t = \arg\min_{\vartheta \in \mR^k} \sum_{i=1}^{n} \frac{1}{\sigma^2}(y_i -\vartheta^\top z_i)^2 + \lambda \norm{\vartheta}^2_{\bS}
	\end{equation}
	has anytime confidence sets for all $t\geq 0$ 
	\begin{equation}
		\pP\left( \norm{\bC\theta - \bC\hat{\vartheta}_t}_{\bOmega_\lambda} \geq 1 + 2\log\left(\frac{1}{\delta}\frac{\det(\bOmega_{t,\lambda})}{\det(\tilde{\bV}_0)}  \right)+ \sum_{i=1}^{t} \norm{\bJ_i}_{{\mV}_0^{-1}}	\right) \leq \delta
	\end{equation}
	where $\delta \in (0,1)$, and $\bOmega_{t,\lambda} = \frac{\bZ_t^\top \bZ_t}{\sigma^2} +  \lambda \bS$, where $\bS= (\bC \mV_0 \bC^\top )^{-1}$
\end{theorem}
Notice that if $\bJ$ is small in the Frobenius norm, the confidence sets improve. In fact, they depend only on the norm. 
\begin{proof}
	\begin{eqnarray*}
		\norm{\bC\theta - \bC\hat{\vartheta}}^2_{\bOmega_\lambda} & = & \norm{\bC\theta - (\frac{1}{\sigma^2}\bZ^\top \bZ  +  \lambda\bS )^{-1}\bZ^\top (\bX \theta + \epsilon)}^2_{\bOmega_\lambda} \\
		& = & \norm{\bC\theta - \bOmega_{\lambda}^{-1}\bZ^\top ((\bZ\bC + \bJ) \theta + \epsilon)}^2_{\bOmega_\lambda} \\
		& = & \norm{\bC\theta - \bOmega_\lambda^{-1} \frac{1}{\sigma^2} \bZ^\top \bZ \bC\theta  + \frac{1}{\sigma^2}\bOmega_\lambda^{-1}\bZ^\top \bJ\theta + \frac{1}{\sigma^2}\bOmega_\lambda^{-1}\bZ^\top \epsilon}^2_{\bOmega_\lambda} \\
	& = & \norm{  \bOmega_\lambda^{-1}(  \bOmega_\lambda - \frac{1}{\sigma^2}\bZ^\top \bZ )\bC\theta  + \frac{1}{\sigma^2}\bOmega_\lambda^{-1}\bZ^\top \bJ\theta + \frac{1}{\sigma^2}\bOmega_\lambda^{-1}\bZ^\top \epsilon}^2_{\bOmega_\lambda} \\
		& = & \norm{  \bOmega_\lambda^{-1}\lambda \bS \bC\theta  + \frac{1}{\sigma^2}\bOmega_\lambda^{-1}\bZ^\top \bJ\theta + \frac{1}{\sigma^2}\bOmega_\lambda^{-1}\bZ^\top \epsilon}^2_{\bOmega_\lambda} \\
		& \leq & \underbrace{\norm{  \bOmega_\lambda^{-1}  \lambda\bS \bC\theta}^2_{\bOmega_\lambda}}_{\text{bias}}  +  \underbrace{\norm{\frac{1}{\sigma^2}\bOmega_\lambda^{-1}\bZ^\top \bJ\theta}^2_{\bOmega_\lambda}}_{\text{self-normalized bias}}  +  \underbrace{\norm{\frac{1}{\sigma^2}\bOmega_\lambda^{-1}\bZ^\top \epsilon}^2_{\bOmega_\lambda}}_{\text{self-normalized noise}} \\
	\end{eqnarray*}
Let us now analyze each term separately. The self-normalized terms can be analyzed using a classical technique with a proper choice of mixture distribution as we show in the proof of the Theorem \ref{thm:adapt_design_ridge_general}. In this case it is a normal distribution $\mN(0, (\lambda \bS)^{-1})$. 

The first term can be shown to be bounded by,
\begin{eqnarray}
\norm{ \lambda \bOmega_\lambda^{-1} \bS \bC\theta}_{\bOmega_\lambda}^2 & = &  \norm{  \lambda \bOmega_\lambda^{-1}(   \bS^{}(\bC \mV_0 \bC^\top) \bS^{} )\bC\theta}_{\bOmega_\lambda}^2 \\
& = & \lambda^2 \theta^\top \bC^\top  {\bS} \bOmega_\lambda^{-1} {\bS} \bC \theta \\
& \leq & \lambda^2 \theta^\top\bC^\top  \bS (\lambda \bS)^{-1} \bS  \bC \theta  \\
& \leq & \lambda \theta^\top\bC^\top (\bC\mV_0 \bC^\top)^{-1}  \bC \theta  \\
& \leq & \lambda \theta^\top  \mV_0 \theta \leq 1
\end{eqnarray}

They can be analyzed in the same spirit as in \citet{Mutny2020} novel term
\begin{eqnarray}
	\norm{\frac{1}{\sigma^2}\bOmega_\lambda^{-1}\bZ^\top \bJ\theta}^2_{\bOmega_\lambda} & = & \frac{1}{\sigma^2}\theta \bJ^\top \bZ \bOmega^{-1}_\lambda \bZ^\top \bJ \theta \\
	& = & \theta \bJ	^\top \frac{\bZ}{\sigma} (\frac{\bZ^\top \bZ}{\sigma^2} + \tilde{\bV}_0)^{-1} \frac{\bZ^\top}{\sigma} \bJ \theta \\
	& \leq & \theta \bJ	^\top \bZ (\bZ^\top \bZ)^{\dagger} \bZ^\top \bJ \theta \\
	& \leq & \theta \bJ	^\top \bJ \theta = \sum_{i=1}^{t} (j_i^\top \theta)^2 \leq \sum_{i=1}^{t} \norm{j_i}_{{\mV}_0^{-1}}^2 \frac{1}{\lambda}	
\end{eqnarray}
\end{proof}

\section{Algorithms}\label{app:algorithms} 
Now we will briefly discuss algorithms that construct designs $\bX$, and allocations over them $\eta$, which lead to low error either in expectation or with high probability - using $A$ or $E$ design, respectively. Depending on the aim and estimator, different algorithms might be preferable. For comprehensive reviews please refer to \cite{Todd2016}, \cite{Pukelsheim2006},\citep{Fedorov1997} for optimization algorithms, and \citep{Allen-Zhu2017}, \citep{Camilleri2021} for rounding techniques. 
	\subsection{Greedy selection}
	A first idea is to greedily maximize the scalarized information matrix with the update rule
	$\eta_{t+1} = \frac{t}{t+1}\eta_t + \frac{1}{1+t}\delta_t$,
	\[ \delta_t = \argmax_{x \in \mX} f\left(\bW_{\lambda}\left(\frac{t}{t+1}\eta_t + \frac{1}{t+1}\delta_x\right)\right)\]
	where $f$ refers to the scalarization and $\delta_x$ to the indicator of the discrete measure corresponding to the feature $\Phi(x)$. Notice that while stated in form of allocations, due to the form of the update rule $t\eta_t$ is always an integer.
	
	Surprisingly, this algorithm can fail with the interpolation estimator, where adding a new row to $\bX$, which does not lie in the kernel of $\bC$ increases the variance. This is an artifact of the pseudo-inverse, but it demonstrates that the algorithm is not universal -- hence we suggest using it with the regularized estimator only.
	\subsection{Convex optimization} \looseness=-1
	Alternatively, one can first select a design space $\bX$ supported on finitely many queries such that if the optimal design is supported on it, it leads to a desired low bias. After that, we optimize the allocation $\eta \in \Delta^n$ over it. This has the advantage that {\em a)} we can bound the query complexity to reach $\epsilon$ of learning as in Proposition \ref{prop:query-complexity}, and {\em b)} we can provably achieve optimality with convex optimization, in contrast to the greedy heuristic.
	
	We look for an allocation using convex optimization methods as in \eqref{eq:opt} or more generally,
	$\max_{\eta\in \Delta^n} f(\bW_\circ(\bD(\eta)^{1/2}\bX))$ where $ \circ \in \{\dagger,\lambda\}$.
	There are three possible ways to initialize the algorithm with $\bX$: make an ansatz, greedily reduce bias first, or use a modification of the random projection initialization of \citet{Betke1992} described below. 
	\paragraph{Mirror-descent and Regularity}
	\looseness -1 The objective above (or \eqref{eq:opt}) would be classically solved via Frank-Wolf algorithm \citep{Todd2016}, or a mirror descent algorithm \citep{Beck2003} \footnote{Mirror descent is known as multiplicative algorithm in the experimental design literature \citep{Silvey1978}.}, which starts with the whole support, and reduces the weight of some of the queries $\eta_i$, as
	\[ \eta_{t+1} = \eta_t \frac{ \exp(-s_t \nabla f(\bW(\eta_t)) ) }{\sum_i \exp(-s_t \nabla_i f(\bW(\eta_t)) ) }\]
	where $s_t$ is the stepsize; if $s_t \propto \sqrt{t}^{-1}$, the convergence is guaranteed. Both $E$ and $A$-design objectives are concave as they are related by linear transform $\bC$ from the classical objectives, which are known to be concave \citep{Pukelsheim2006}.
	
	paragraph{Initialization via random projections}
	This algorithm is inspired by volume algorithm of \citet{Betke1992}. The algorithm proceeds by picking a random vector $c_0$ in the span of $\bC^\top$ and picking two points $\bar{z}, \underline{z} = \arg\max_x c^\top x, \arg\min_{x} c^\top x$. Subsequently, we pick another $c_1$ which is orthogonal to $\underline{z} - \bar{z}$ and still in span of $\bC^\top$. We repeat this procedure $k$ times. Should this not generate a sufficiently accurate starting point with desired bias (as measured by Def.~\ref{def:approx}), we repeat this procedure until such design is obtained.
	
	\subsection{SDP reformulations of the objectives}\label{app:SDP}
	We will now show that both $A$ and $E$ designs have an associated semi-definite reformulations which are possible if $\dim(\mH_\kappa)=m$ \footnote{We would like to thank Stephen Wright of University of Wisconsin for suggesting to revisit this idea.}. One should bear in mind that these are possible only if the $\bC \theta$ is estimable in the sense of Def. \ref{def:estimability}, Hence their might not be utilizable in many instances that this paper studies, and rather apply in cases classical experiment design. Nevertheless they do not appear in literature to the best of our knowledge. 
	
	On top of that these problems are more difficult to solve that lets say the alternative algorithm with mirror descent, however with the use of off-the shelf solvers for SDP problems such as cvxpy, they can be conveniently implemented. These refomrmulations are inspired by classical reformulations due to \citet{Boyd2004}, which show these for estimating $\theta$ directly. The key ingredient is to make use of generalized Shur-complement theorem of \citet{Zhang2011}.
	
	\paragraph{A-design} objective is $\max_{\eta \in \Delta_n} 1/\Tr(\bW(\eta)^{-1})$  can be represented as semi-definite program:
	\begin{eqnarray*}
		&\min_{u\geq 0, \eta \in \Delta_n} &u_1 + u_2 + \dots + u_k \\
		 &\text{subject to} &~ \begin{pmatrix}
			 	\bV(\eta) & \bC^\top e_i \\
			 	e_i^\top \bC & u_i  \\
			 \end{pmatrix} \succeq 0 ~ \text{ for all} ~ i \in [k]
	\end{eqnarray*}
	\paragraph{E-design} objective is $\max_{\eta \in \Delta_n}\lambda_{\min}(\bW(\eta))$, can be again solved using semi-definite programming as:
	\begin{eqnarray*}
		&\min_{t\in \mR, \eta \in \Delta_n} &t \\
	&	 \text{subject to} &~ \begin{pmatrix}
			\bV(\eta) & \bC^\top  \\
			\bC & t \bI_{k\times k}  \\
		\end{pmatrix} \succeq 0
\end{eqnarray*}
We will show how to apply the generalized for the first objective (A-design) only. The second reformulation follows analogously. Notice that the objective $\max_{\eta} 1/\Tr(\bW^{-1})$ is equivalent to $\min_{\eta} \Tr(\bW^{-1})$, and $\bW^{-1} = \bC \bV(\eta)^{\dagger}\bC^\top$. Generalized Shur-complement lemma of \citet{Zhang2011} states that $\bC\bV(\eta)^\dagger \bC^\top  \preceq \bLambda$ with the projection operator lying in the null space of $\bC$, i.e. $(\bI - \bV(\eta)^\dagger\bV(\eta))\bC^\top=\mathbf{0}$, is equivalent to 
\[\begin{pmatrix}
	\bV(\eta) & \bC^\top \\
	\bC & \bLambda  \\
\end{pmatrix} \succeq 0. \]
The second condition is satisfied as long as the problem is estimable, in other words, there exists $\bL$ s.t. $\bC = \bL \bX$. 
Hence minimizing the trace of $\bLambda$, we can minimize the trace of the information matrix. As the trace depends only on the diagonal elements of $\bLambda$, we can choose it to be diagonal, i.e., $\bLambda = \operatorname{Diag}(u)$. 
\subsection{ Query complexity and Optimization}

The objective in \eqref{eq:opt} was stated in the kernelized form which is only valid if the RKHS is finite-dimensional. We state here for completeness the version which operates in the kernelized setting where all the operators can be inverted (and calculated explicitly using only the kernel $\kappa$ evaluation). 

\begin{equation}
	\eta^* = \argmax_{\eta \in \Delta^n}\lambda_{\min} \left( \bC\mV_0^{-1}\bX^\top (\bD(\eta)\bX \mV_0^{-1} \bX^\top\bD(\eta))^{-2\dagger}\bX\mV_0^{-1}\bC^\top \right)^{-1}.
\end{equation}

\section{Geometry}\label{app:geometry} 
We saw in Section \ref{sec:complexity} that properties of $\{\Phi(x): x \in \mS \}$ enter into consideration when we want to bound the total number of queries required to reach $\epsilon$ accuracy with high probability such that we can learn $\bC\theta$. The dependence enters as the smallest eigenvalue $\lambda_{\min}$ of the information matrix $\bW$. In this section, we relate the eigenvalue $\lambda_{\min}(\bW)$ directly to the geometry of the above set using seminal results about E-experimental design from \citet{Pukelsheim1993}.

\subsection{Set Width}\label{app:algorithm:width}
The geometrical property we are interested in \emph{width of a set} that we define below. However, first, we want to make a general note that the formal results of our theorems hold for symmetrized sets $\{\Phi(x)|x\in \mS\}$. This is without the loss of generality, since, an optimal design on symmetrized set and a non-symmetrized set is equivalent. This can be seen by considering $\bW_1(\eta)$ defined on symmetrized set equates the one of non-symmetrized set $\bW_2(\eta)$ due to the fact evaluations $\Phi(x)$ enter only as outer products $\bW_1(\eta) = r(\sum_{i=1}^n \eta_i (\Phi(x_i)\Phi(x_i)^\top))$, where the sign clearly does not change anything in terms of the information matrix, and $r$ correspond to further operations to define $\bW$ properly (i.e. projection and inverse).

\begin{definition}[Width] Let $\mD$ be a convex set in $\mH$, then the width of it is defined as,
	\begin{equation}
		\operatorname{width}(\mD) = \min_{\theta \in \mH} \max_{x,y\in \mD} \theta^\top(x-y)
	\end{equation}
	where $\theta$ is the unit norm.
	
	\emph{Width in the span of $\bC$} is defined to be,
	\begin{equation}
		\operatorname{width}(\mD,\bC) = \min_{\theta = \bC^\top\alpha ~ \text{s.t.} ~ \alpha \in \mR^k} \max_{x,y\in \mD} \theta^\top(x-y)
	\end{equation}
	where $\alpha$ is the unit norm. 
\end{definition}
\begin{definition}[Gauge norm]\label{def:gauge}
	Let $\mD$ be a set of points in $\mH$, then the gauge norm of $\mD$ is, 
	\begin{equation}\label{eq:gauge-norm}
	 \rho_{\mS}(x) = \inf_{r\geq 0}\{r : x \in r\conv(\mD) \} 
	\end{equation}
\end{definition}
\begin{theorem}[Thm. 2.2 in \citep{Pukelsheim1993}]\label{thm:pukels}
	Let $\bW$ be any the information matrix for $\bC\theta$, and $z \in \mR^k$ non-zero then
\[	\lambda_{\min}(\bW) \leq \frac{\norm{z}^2_2}{\rho(\bC^\top z)^2}. \]
It holds with \underline{equality} if $\bW$ has smallest eigenvalue of multiplicity one and $z$ is the associated eigenvector. 
\end{theorem}
The treatment without multiplicity is somewhat more complicated and can be found in. 
\begin{proposition}
	Under assumption of Thm.~\ref{thm:pukels}, no multiplicty, $z$ being eigenvector of $\bW$, let $\bC:\mR^m \rightarrow \mR^p$, $\mD$ be a symmetric convex set in $\mH$, then 
\[ \lambda_{\min}(\bW) = \frac{\norm{z}^2_2}{\rho(\bC^\top z)^2}  \geq \frac{1}{\operatorname{width}(\mS, \bC)^2}\]
\end{proposition}
\begin{proof}
	Due to homogenity of the gauge norm, $\rho(\bC^\top z \frac{\norm{z}_2}{\norm{z}_2} ) = \norm{z}^2(\rho \bC^\top \hat{z}) $, where $\hat{z}$ is a unit vector. Thus, the optimization can be restated as, 
	\[   \frac{\norm{z}^2_2}{\rho(\bC^\top z)^2} =  \frac{1.}{\rho(\bC^\top \hat{z})^2} = \frac{1}{\operatorname{width}(\tilde{\mD})^2} \geq \frac{1}{\operatorname{width}(\mD, \bC^\top\hat{z})^2}\]
where in the second to last step we use the fact that the set is symmetric and $\hat{z}$ is unit eigenvector. The last inequality follows by minimizing over all possible unit direction as int the definition of width. The set $\tilde{\mD} \subset \mH$ s.t. $\mD=\{f \in \mH_k | f = \bC^\top \hat{z} \}$. Notice that this set lies in $1$-dimension subspace, and is directly proportional to the gauge-norm.
\end{proof}

Given the above theorem, we can calculate the worst-case value of parameter $\frac{1}{\lambda_{\min}(\bW)}$ that bounds the complexity if the geometry of the set can be understood easily. Alternatively, we can resort to direct calculation. For example consider the gradient design problem, in which we know that for finite difference problem can be bounded by $dh + O(h)$, where $h$ is the stepsize. 

\begin{proposition}
Under the assumption of the model \eqref{eq:model}, if $\bC = \nabla_x\Phi(x)$ at $x \in \mR^d$, $\Phi(x)$ is the evaluation functional, and the design space $\bX$ contains rows $\{\Phi(x\pm he_i\}_{i=1}^d$ where $e_i$ corresponds to the unit vectors in $\mR^d$ and $h$ is the step-size, then
\begin{equation}
    \frac{1}{\lambda_{\min}(\bW(\eta)} \leq dh + O(h^2),
\end{equation}where $\eta$ corresponds to the design supported on all points with equal weight. 
\end{proposition}
\begin{proof}
The inverse of the information matrix is, 
    \begin{eqnarray*}
        \bW_\dagger^{-1} & = & \nabla_x\Phi(x)\left( \sum_{i=1}^{d}\eta_i(\Phi(x + he_i)\Phi(x + he_i)^\top + \Phi(x - he_i)\Phi(x - he_i)^\top) \right)^{-1}\nabla_x\Phi(x)^\top\\
        & = &  \nabla_x\Phi(x)\left( \sum_{i=1}^{d}\eta_i(2\Phi(x)\Phi(x)^\top + h\nabla_x\Phi(x)^\top e_ie_i^\top\nabla_x\Phi(x)  + O(h^2)\right)^{-1}\nabla_x\Phi(x)^\top \\
        & \preceq & \nabla_x\Phi(x)\left( h\nabla_x\Phi(x)^\top \bD(\eta) \nabla_x\Phi(x)  + O(h^2)\right)^{-1}\nabla_x\Phi(x)^\top \\
        & = & h \nabla_x\Phi(x)\left( \nabla_x\Phi(x)^\top \bD(\eta) \nabla_x\Phi(x)  + O(h^2)\right)^{-1}\nabla_x\Phi(x)^\top \\
        & = & dh \nabla_x\Phi(x)\left( \nabla_x\Phi(x)^\top \nabla_x\Phi(x)  + O(h^2)\right)^{-1}\nabla_x\Phi(x)^\top \\
        & \preceq & dh \bI + O(h^2)
    \end{eqnarray*}
    In the first step we used Taylor's theorem and keep track of the order components. In the second we used that $\Phi(x)\Phi(x)^\top$ is psd, later we used the definition of the design and the properties of the projection matrix. Lastly,
    \[ \frac{1}{\lambda_{\min}(\bW_\dagger)} = \lambda_{max}(\bW_\dagger^{-1}) \leq dh + O(h^2),\]
    which finishes the proof.
\end{proof}

\begin{corollary}[Query complexity]\label{prop:query-complexity}
Under assumption of Prop. \ref{thm:fixed_design_pseudo} where $\eta^*$ is the optimum to Eq. \ref{eq:opt}, it holds that
		\[ \norm{\bC(\hat{\theta}_\dagger - \theta)}_2 \leq \sqrt{\frac{ \lambda_{\min}(\bW_{\dagger}(\eta^*))^{-1}}{T} }\left(\sigma\sqrt{\xi(\delta)} + \frac{\nu}{\sqrt{\lambda}} \right).\]
with probability $1-\delta$.
\end{corollary}
\begin{proof}
\begin{eqnarray}
\norm{\bC(\hat{\theta}_\dagger - \theta)}_2^2 & \leq & \frac{1}{\lambda_{\min}(\bW_{\dagger}(\eta^*)} \norm{\bC(\hat{\theta}_\dagger - \theta)}_{\bW_\dagger(\eta^*)} \\
& = & \frac{1}{\lambda_{\min}(\bW_{\dagger}(\eta^*)} (\sigma^2 \xi(\delta) +
\frac{\nu^2}{\sqrt{\lambda}})
\end{eqnarray}
Now the $\bW_\dagger$ nor $\nu^2$ do not depend on $\sigma^2$, however evaluating multiple times the same measurements Namely $T$ times, reduces the variance $\sigma^2$ by $1/T$. After taking this into consideration and the square root it finishes the proof. 
\end{proof}

\section{Extra Examples}\label{app:examples}\label{app:extra}
\begin{figure}
	\centering
	\includegraphics[width = 1\textwidth]{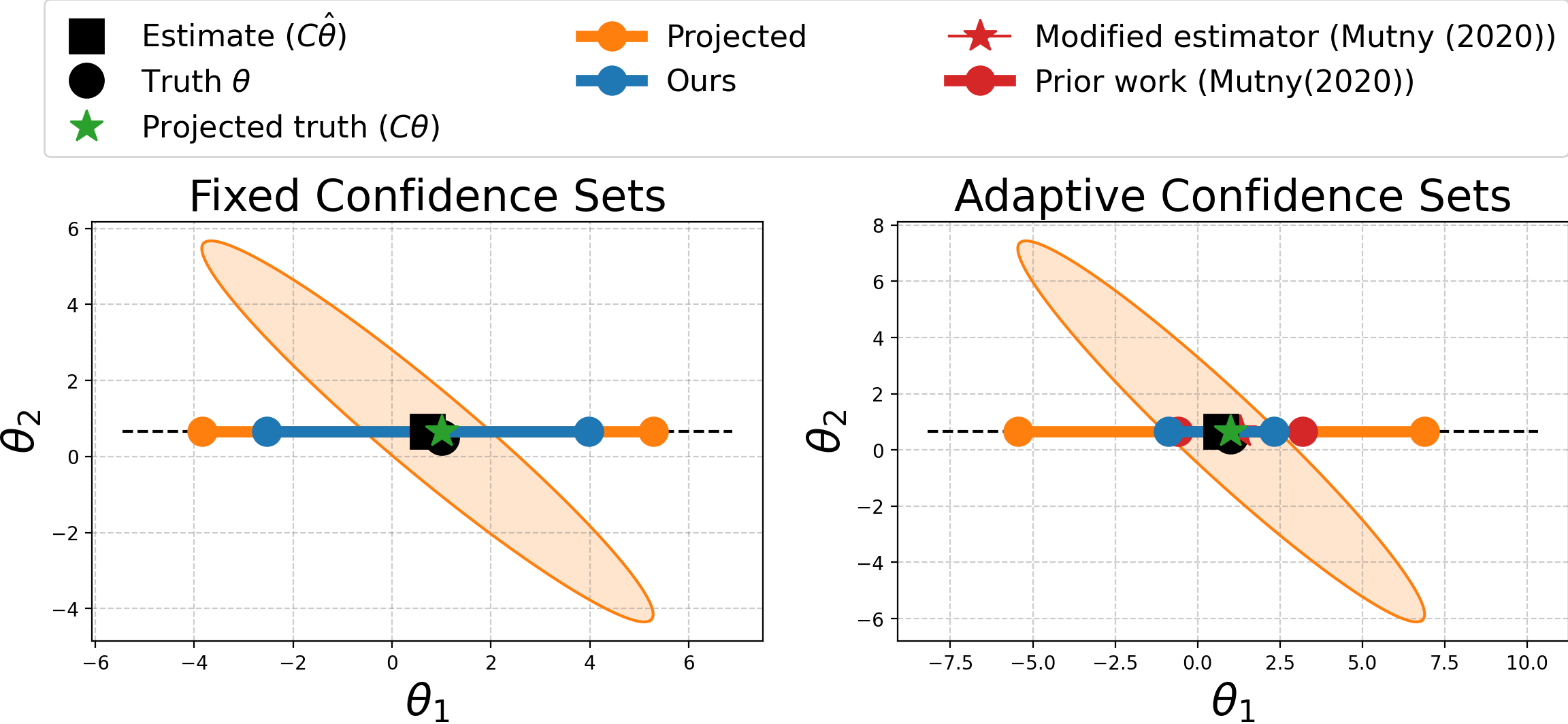}
	\caption{An example of confidence ellipsoids and intervals in two dimensions. You can see that in both regimes: fixed and adaptive. Our confidence sets (Ours) are the tightest for the estimation of the linear projection of $\bC\theta$ value of the parameter $\theta = (1,1)$, in this case $\bC = (1,0)$. On the left, we see fixed confidence sets while on the right are adaptive both with $T=50$. We compare with \citep{Mutny2020} and associated confidence sets, they are larger for this example (and grow with $T$); also the estimator is different as we delineate in Sec.~\ref{app:relation-prior}. }
	\label{fig:ellipse}
\end{figure}

In this section, we provide additional applications of our framework that deserve a mention, but before we do, we want to bring attention to Figure \ref{fig:ellipse}, where we show a comparison of our confidence sets with prior work and projected ones, similar to Figure \ref{fig:ellipse-main}. As a refresher, recall that
\begin{eqnarray*}
	\bL: \mR^n \rightarrow \mR^p \\
	\bX: \mH_k \rightarrow \mR^n \\
	\bC: \mH_k \rightarrow \mR^p.
\end{eqnarray*}

\subsection{Integral maps}

\paragraph{Quadrature} An integral is a special linear operator on the function space, \[
\bC = \int q(x)\Phi(x)^\top \cdot dx =\int q(x)\Phi(x)dx^\top\cdot,\] where $p = 1$. The approximate estimability condition can be written in a concise form \[\norm{\bC - \bL\bX}_k = \norm{\int (q(x)-\sum_{y\in \mS}l(y)\delta(x-y))\Phi(x)dx}_k,\] where $l(y) = \bL_y$, and $y\in \mS$ which itself is equal to well-known integral distance $\operatorname{MMD}_{\mH_\kappa}(q, \sum_{y\in \mD}l(y)\delta(\cdot-y))$ referred to as Maximum Mean Discrepancy (MMD) multiplied by $B$. It measures the worst case integration error on the class of functions $\theta \in \mH_\kappa$ s.t. $\norm{\theta}_k \leq 1$ with nodes $x \in \mS$ and weights $l(x)$. The method which minimizes this quantity is sometimes referred to as Bayesian Quadrature \citep{Huszar2012}. Note that minimization of this quantity would correspond to the interpolation estimator. 

\paragraph{Fourier Spectra: Low-pass Filters}\label{sec:example:spectra}
Suppose we are interested in approximately learning the Fourier spectrum of $\theta$. This might be especially interesting if $\theta$ is composed of two signals one with  specific low-frequency bands and the rest occurs only elsewhere and is of no interest. Let $\{\omega_i\}_{i=1}^p$ be the frequencies of interest, then the rows of $\bC$ are $\bC_j = \int_\infty^\infty \exp(i\omega_jx) \Phi(x)^\top \cdot dx$, which can be discretized or evaluated in closed form depending on the kernelized space.

\paragraph{Transductive Risk}\label{sec:example:error}
If the dimension of the response vector is very large much larger than the number of elements in the unlabeled set, which is a common appearance in deep learning or high dimensional statistics; there is still hope to have provable bounded error on selected important candidates. As opossed to study the full risk $ \mE_{\epsilon}[\mE_{x \sim \rho}[(x^\top(\theta-\hat{\theta}))^2]]$ with data distribution $\rho$, we can optimize the risk on a set of candidates $\mE_{\epsilon}[\sum_{i}^{p}[(q_i^\top(\theta-\hat{\theta}))^2]]$, which becomes $\Tr(\bQ\bV^\dagger\bQ)$, where $\bQ$ contains the rows of $k$ are the candidates.

\subsection{Solution to Contamination (Low pass filter)}\label{sec:example:contamination}\looseness -1
Often in estimation, the signal of interested is corrupted by another signal that we are not interested in. This is sometimes referred to as \emph{contamination}  \citep{Fedorov1997}. In particular, consider the following model
$ \mE[y]  = \begin{pmatrix}
	\alpha &\beta
\end{pmatrix}^\top \begin{pmatrix}
	\phi(x) & \psi(x)
\end{pmatrix}$,
where we want to infer $\alpha$ only. Note that this is a special case of our framework, where the  linear functional $\bC = \bI_S$ zeros out the variables $\beta$ in $\theta=(\alpha, \beta)$. This problem is well-defined only if the spans of $\{\phi(x)\}_{x\in \mS}$ and $\{\psi(x)\}_{x\in \mS}$ are not contained in each other, which we neglect for our purposes here. 

For illustration, consider a linear trend contaminated with function $f$, which has major frequency components that are {\em known}, i.e., $y = \alpha^\top x + f(x) + \epsilon$. We can stack these frequencies in an evaluation functional $\psi(x,\{\omega\}_{i=1}^m)$, and express the problem in the form above. In Fig.~\ref{fig:contamination}, we report the MSE for estimation of $\alpha$ with $f$ which has frequency components $\{\pi l,\pi e l | l \in [16]\}$ weighted with $1/l^2$. The contamination-aware design tries to query points that eliminate the effect of $f$ and leads to lower MSE for the same number of observations and performs much better in expectation than either random or full designs -- an optimal design that learns both $\alpha$ and $\beta$. For this design, we chose to sample greedily first to reduce bias and then optimized the weights of each data point to achieve optimality. 

\subsection{Learning ODE solutions}
\begin{figure}
	\centering
\begin{subfigure}[t]{0.45\textwidth}
	\includegraphics[width=\textwidth]{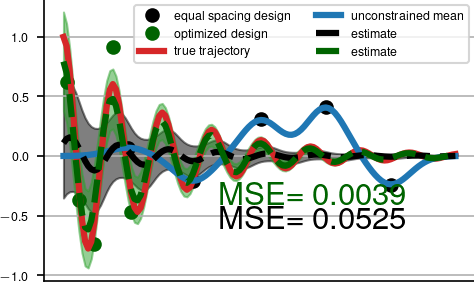}
	\caption{An estimate of the trajectory of a damped harmonic oscillator. The source of unknown here is due to the unknown initial conditions. Notice that with the optimized design we can achieve a much better fit. Also, the unconstrained fit with the squared exponential kernel is shown in blue. Below we also see the reduction in MSE for estimating this trajectory with the two designs. }
	\label{fig:odeint}
    \end{subfigure}
~
\begin{subfigure}[t]{0.45\textwidth}
	\includegraphics[width=\textwidth]{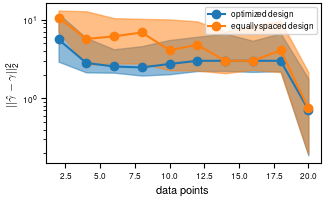}
	\caption{MLE error of estimating $\gamma$ in the pharmacokinetic study. We see that with the optimized design we are consistently better in terms of MSE than equally spaced design. Error bars with 20 repetitions included. The y-axis denotes the number of data samples (draws of blood) required. Notice the logarithmic scale.}
	\label{fig:pharm}
\end{subfigure}
\caption{Further experiments showing application in ODE trajectory and parameter estimation. }
\label{fig:odeint-pharm}
\end{figure}

 Consider an example of damped harmonic oscillator, $u(t)'' + \kappa u(t)' + u(t) = 0 ~ \text{for} ~ t \in [t_0, t_1]$. Adopting the above procedure and using the squared exponential kernel to embed trajectories \footnote{The nullspace of $\bZ$ forms its own Hilbert space that spans the space of trajectories, however for estimation convenience it is sometimes easier to work with a larger space and then project.} , we show in Figure \ref{fig:odeint} inference with the shape constraint using $A$-optimal design as well as equally space design, we see the resultant confidence bands and fit are much better for optimized design. We assume that $\kappa \in (\kappa_-,\kappa_+)$ and give a union of confidence sets as in the robust design case. The optimal design then corresponds to the unknown null subspace $\bC$ and its variance part is equal to $\sup_{\kappa \in [\kappa_-, \kappa_+]}\Tr(\bC_\kappa\bV^\dagger\bC_\kappa^\top )$.

\subsection{Learning PDE solutions}
In a similar fashion as the linear ODE feature, a linear constraint so do linear PDEs. We demonstrate this on a sensor placement problem, where consider heat equation in two dimensions with $u(t,x)$ with varying diffusion coefficient $c(x)$.
\begin{eqnarray*}
	\partial_t u - c(x)\partial_{xx} u  = 0  \quad \text{in} \quad (0,T)\times \Omega \\
	u(0,x) = \braket{\theta,\Phi(x)}\quad \text{for } \quad  x \in \Omega,  z \in \partial \Omega.
\end{eqnarray*}
We assume that the point value of initial conditions $u(0,x)$ can be measured by placing sensors along $x$. The goal is to place these sensors such that after elapsed time $T$, $u(T)$ can be inferred with the lowest error. This is the forward formulation of the inverse problem in \citet{Leykekhman2020}. We assume that $\theta$ is a bounded member of RKHS due to the squared exponential kernel. In this case we are not interested in the full nullspace of $\bN$ as in the previous example, instead we focus on the range space of $\bC= \Phi(x,T)(\bN^\top \bN)$ corresponding to information at $\Phi(x,T)$.

\subsection{Sequential Design: Estimating CVar}\label{app:example:risk}
Often the objective we are trying to estimate is inherently stochastic such as the following model:
\begin{equation}
	\mE_{z \sim W(x)}[\rho(\Phi(x,z)^\top \theta)].
\end{equation}
In particular one can be interested in $x$ s.t. $\arg\sup_{x}	\mE_{z \sim W}[\Phi(x,z)^\top \theta]$. This is analyzed in sequential experiment design optimization with risk measures, where $\rho$ is the risk measure (see \citep{Cakmak2020} and \citep{Agrell2020}). It is often assumed that $W$ cannot be evaluated explicitly, only sampled from with the evaluation oracle $y_i = \Phi(x_i, z_i)^\top + \epsilon_i$. We focus on a problem where we want to learn the risk of the actions $q \in Q, |Q|=k$, $\mE_{z \sim W}[\rho(\Phi(q,z)^\top \theta)]$ instead of minimization and $\rho$ is CVar risk measure. With a fixed value $\theta$, this is a linear operator $\bC_\theta$. Since $\theta$ is unknown, we adopt a sequential procedure, where we solve for the risk $C_{\hat{\theta}}$, where $\hat{\theta}$ is sampled uniformly from its confidence set.

\begin{figure*}
	\centering
	\begin{subfigure}[t]{0.45\textwidth}
		\includegraphics[width=\textwidth]{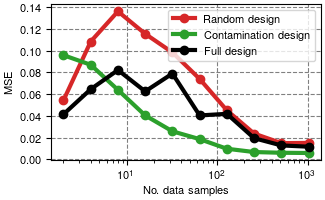}				\caption{Mean squared error of the estimated linear trend. We see that contamination-aware design improves over full design as well as randomly selected design. }
	\end{subfigure}
~
	\begin{subfigure}[t]{0.45\textwidth}
		\includegraphics[width=\textwidth]{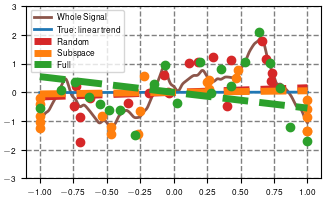}
		\caption{Example of the signal in brown with design points as well as linear trend estimates in color-coding. Full denotes joint estimation of brown and blue signal. Random designates a random design. }
	\end{subfigure}
\caption{Statistical contamination: a numerical experiment}
\label{fig:contamination}
\end{figure*}

\section{Experiments: Details}\label{app:experiments}

\subsection{Gradient Estimation: Details}
With gradient estimation we are interested in $\nabla_x \Phi(x)$ of an element $\theta \in \mH_\kappa$. The example generated in Fig.~\ref{fig:grads} is in 2 dimensions, where we approximate $\Phi(x)$ of squared exponential kernel with lengthscale $l = 0.1$ with Quadrature Fourier Features of \citet{Mutny2018b} for the convenience of optimization. We consider a slightly off-set finite difference design, with the following elements, 
\[ \{\Phi(x), \Phi(x + he_1), \Phi(x +2he_1), \Phi(x-he_1), \Phi(x-he_2)\}. \]
We calculate the relative-bias $\nu$ due to the Def. \ref{def:approx} and balance it with the budget (as in the Figure). This way we can identify the optimal $h^*$ before even solving for the optimal design. Given the identified $h^*$, we calculate the optimal design with mirror descent solving the objective \eqref{eq:opt}, to get
\[ \eta^* = (0.37,0.09,0.08,0.09,0.38), \]
which is not obvious without understanding the geometry of the problem. The value of $\sigma$ was set to be $0.01$. The total error in Fig.~\ref{fig:grads} is the combination of bias and variance which was calculated using the largest eigenvalue of $\bE(\bL_\dagger)$ as in \eqref{eq:residuals}. This is the error we can guarantee with high probability (not the actual error) which is somewhat lower. 

\subsection{The solution to Contamination (Statistics): Details}
For this example, consider a linear trend contaminated with $f(x)$ which has known major frequency components, i.e., $y = \alpha^\top x + f(x) + \epsilon$. We can concatenate these frequencies, and construct an evaluation operator where $\Psi(x)_i = \cos(\omega_i x)$, and $\Psi(x)_{i+1} = \sin(\omega_i x)$ where $\omega_i$ is one of these frequencies. Then, we express the problem as $y = \theta^\top \Phi(x) =  (\alpha, \beta)^\top (x, \Psi(x, \{\omega_i\}_{i=1}^m) + \epsilon$. Since we are interested only in $\alpha$, the operator $\bC = (1, \mathbf{0})$ selects it. Hence we look for a design (subset of points) which only minimizes the residuals due to estimation of $\alpha$. In fact, such design might indirectly minimize the other residuals as well, but only if this is necessary to reduce the residuals due to $\alpha$. 

The example depicted in Fig.~\ref{fig:contamination} is created using the frequencies $\{\pi l| l \in [16]\}\cup\{\pi e l | l \in [16]$ weighted with $1/l^2$. Such weighting of high frequencies creates contaminating signals that are similar to functions from Mat\'ern kernel spaces \citep{Rasmussen2006}. The contamination-aware design performs much better. 

The values of $\sigma = 0.5$ (relatively high). We compare with full design -- one that minimizes the overall estimation error or in other words, where $\bC$ is the identity. We first use the greedy algorithm to minimize the regularized estimator to obtain initial design space. Then we use convex optimization to find a proper allocation of a fixed budget (varying in the x-axis) to these queries. We again optimize with mirror descent. 

\subsection{Pharmacokinetics: Details}
The pharmacokinetic two-compartment model models two organs. In this case the stomach and blood and the transfer of the medication between them. The model can be mathematically described as 
\begin{eqnarray*}
\frac{d}{dt} c_s(t) & = & -a c_s(t) \\
\frac{d}{dt} c_b(t) & = & bc_s(t)-d c_b(t) 
\end{eqnarray*}
where $c_s(t)$ represent concentration of the medication in the stomach, and $c_b(t)$ represents the concentration in the blood. The goal of pharmacokinetic studies is to identify $(a,b,c)$ from draws of the blood of subjects to e.g. properly assign dosage. 

In comparison to the example of the damped harmonic oscillator, the initial conditions are known -- albeit imprecisely -- what is the true unknown is $\gamma = (a,b,c)$. We make the following consideration, in order to infer $\gamma$ precisely we need to have a precise estimation of the trajectory, hence we inject a slight perturbation to the initial condition and then search for the optimal design that is good for any of the models $\gamma$: robust design. This way with any of the models from $\gamma \in \Gamma$ we will have a good estimation of trajectories and hence a good estimation of $\gamma$ using MLE as we report in Fig.~\ref{fig:pharm}.

The specific parameter used was $\sigma = 0.01$ which corrupted the observations of $c_b(t)$ at chosen times and the space which Embeds the trajectories was chosen to be that of the squared exponential kernel with lengthscale $l = 0.05$. The true parameters were set to be $(5,10,10)$ and the robust set of $\Gamma = ((4,6),(9,11),(9,11))$ for each parameter. We use the greedy algorithm with regularized estimator and $\lambda = 1/2$ to optimize the robust $A$-optimal metric. The resultant set is visualized in Fig.~\ref{fig:pharma}.

\subsection{Control: Details}
\begin{figure}
\centering
	\begin{subfigure}[t]{0.45\textwidth}
		\includegraphics[width=\textwidth]{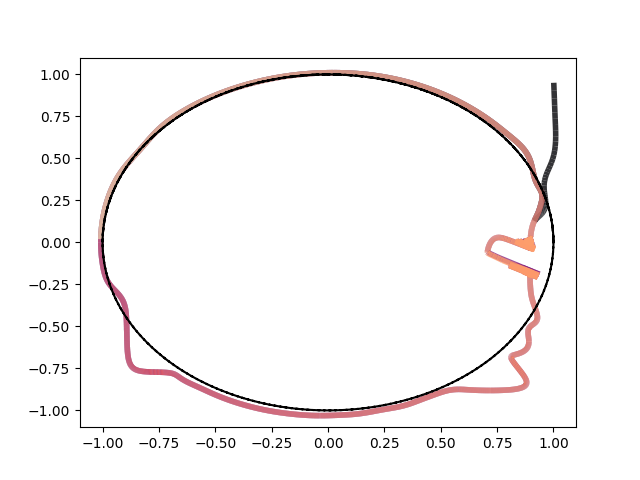}
		\caption{An example of an unstable trajectory around the reference trajectory, started from (1,1) point}
	\end{subfigure}
	~
	\begin{subfigure}[t]{0.45\textwidth}
	\includegraphics[width=\textwidth]{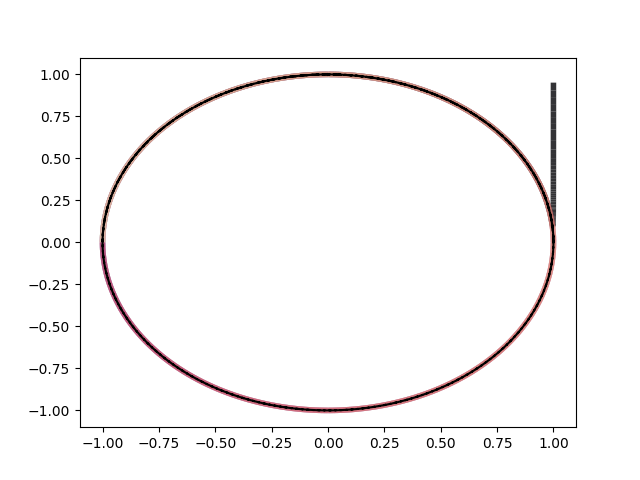}
	\caption{An example of a stable trajectory around the reference trajectory, started from (1,1) point}
	\end{subfigure}
~
	\begin{subfigure}[t]{0.45\textwidth}
	\includegraphics[width=\textwidth]{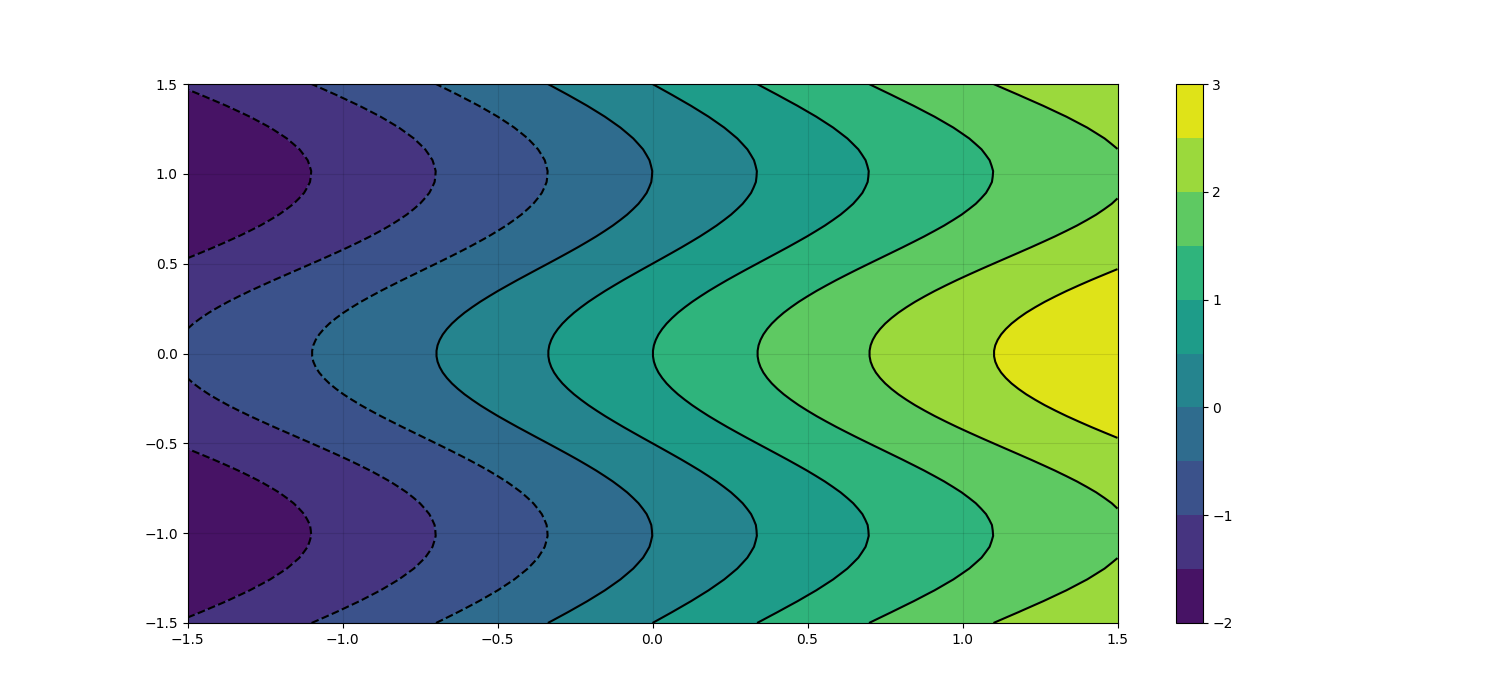}
			\caption{Non-linear dynamics function $f_1(x)$ which governs the first component $x$.}
	\end{subfigure}
	~
	\begin{subfigure}[t]{0.48\textwidth}
	\includegraphics[width=\textwidth]{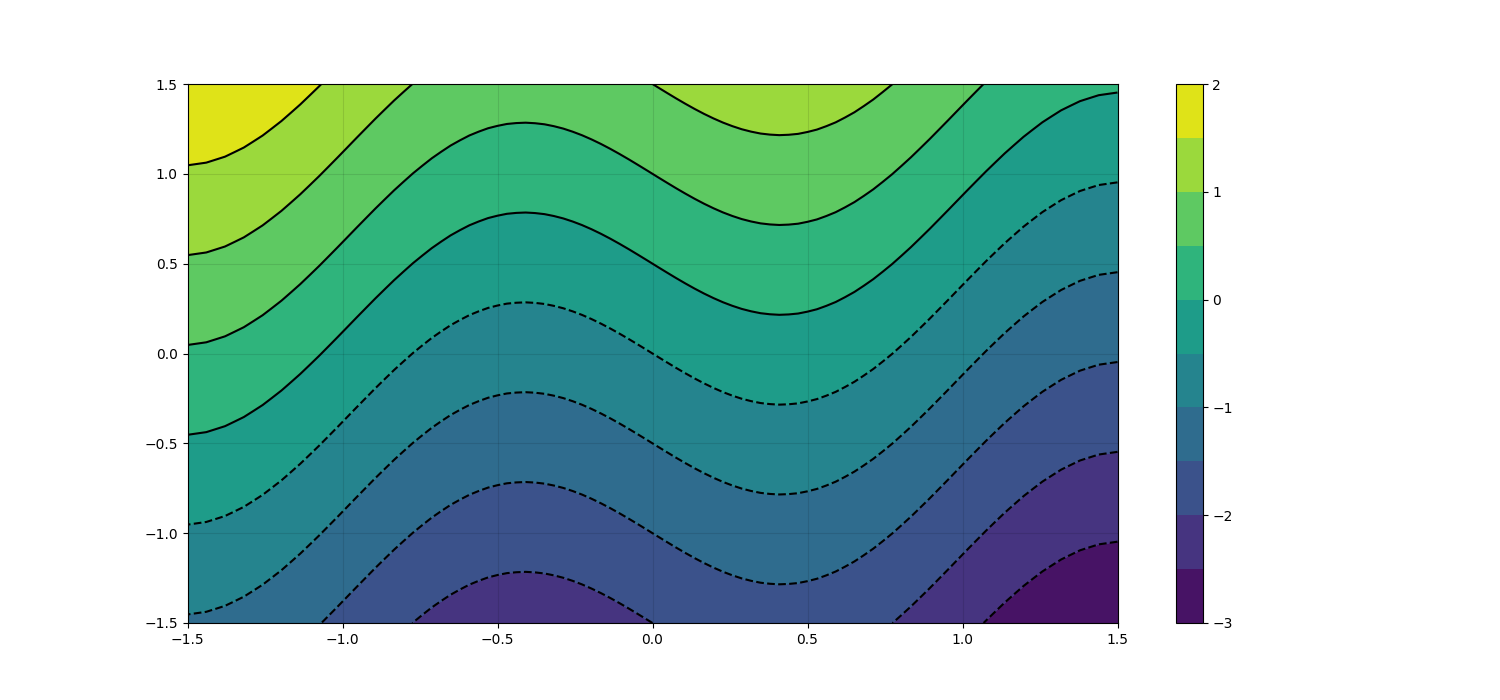}
	\caption{Non-linear dynamics function $f_2(x)$ which governs the second component of $x$.}
\end{subfigure}
\caption{Control Experiment: Additional details. }
\label{fig:control-extra}
\end{figure}

In the last experiment, we essentially model the same problem as in \citet{Lederer2020} as we explain. The only difference in our setup is the use of a gain $K = 200$ instead of $K=15$. With a low gain of $K=15$ one cannot provably stabilize with the given Lyapunov function. With $K=200$ we were able to quickly stabilize the system once enough data points were obtained.  Notice the reference trajectory and trajectory with "bad" and "good" controllers in Fig.~\ref{fig:control-extra}. The driving non-linear dynamics are visualized likewise. 

The experiment is designed to showcase the merit of the newly designed adaptive confidence sets with the stopping rule that stops once the controller has been verified to be stable. We repeated each experiment 10 times and reported the standard quantiles of the total derivative of the Lyapunov function in Fig.~\ref{fig:control}. We always started with the initial set of 10 data points and then proceeded with exploration (adding specific data points) if the supermum of the total derivative was not negative. When it was we stopped. The stopping corresponds to the line going into negative values in the log plot in Fig.~\ref{fig:control}. We considered these four strategies:
\begin{itemize}
    \item \emph{random} - randomly sampling data point from the whole domain [-1.5,1.5].
    \item \emph{random-ref} - randomly sampling data point around the operating region.
    \item \emph{unc} - sampling (greedily) the most uncertain query from the whole domain [-1.5,1.5].
    \item \emph{unc-ref} - sampling (greedily) the most uncertain query from the operating region.
\end{itemize}
\citet{Berkenkamp2016} provides a better solution than uncertainty sampling, which is a special case of the linear functional studies in this work. In this example, this closely follows the \emph{unc-ref} baseline and would create additional clutter. The inclusion of this algorithm would not further validate the benefit of the tighter confidence sets. The operating region is chosen to be a tube around the reference trajectory of width $h = 0.01$. 

The dynamics is modeled using Nyst\"om features ($m = 400$) of the squared exponential kernel with lengthscale $0.25$ and regularized estimator with $\lambda$ estimated experimentally. We selected this with visual inspection since the example is known, but one can adopt the procedure in \citet{Umlauft2018}. The noise std. $\sigma = 0.05$ and number of queries is $T=500$.

\section{Improved regret of linear bandits: Proofs}\label{app:bandits}
To prove the following theorem we require the famed potential lemma which we state for completeness, 
\[\sum_{t=1}^T \norm{x_t}_{\bV_{t}^{-1}} \leq \mO(\sqrt{dT\log(T/d)}).\]
A reference for this result can be found in \citep{Szepesvari2019}. Note that a bounded pay-off vector assumption has been used in the derivation. In general this result can be extended to kernelized bandits by noting that 
\[\sum_{t=1}^T \norm{x_t}_{\bV_{t}^{-1}} = \mO(\sqrt{\gamma_T T}),\]
where $\gamma_T = \log\left(\frac{\det(\bV_t)}{\det(\bI)}\right)$, due to matrix inversion lemma, this can be also expressed in kernelized form, as is usually referred to as information gain \citep{Srinivas2009}.
\begin{theorem}\label{thm:linear-bandits}
    Let $\theta \in \mR^d$ be an unknown pay-off vector, and a set of actions $x \in \mX$ such that $|\mX| \infty$, then the regret of UCB algorithm is no more than
    \begin{equation}
        R_T \leq \mO(\sqrt{dT\log (T(|\mX|+1)/\delta)} )
    \end{equation}
    with probability $1-\delta$. 
\end{theorem}
\begin{proof}[Proof of Theorem \ref{thm:linear-bandits}]
We drop $\Phi$ and use $x$ instead only. Note that due to Lemma \ref{lemma:ordering-adaptive}, we have that \[\norm{x_k^\top(\hat{\theta}-\theta)}_{\bW_{t,k}}^2 \leq \norm{x_k^\top(\hat{\theta}-\theta)}_{\bOmega_{t,k}}^2\leq \beta_t,\] where $\beta_t$ is defined for the information matrix $\bOmega_{t,k}$ in Theorem \ref{thm:adapt_design_ridge_general}. The subscript $k$ designates the dependence on $x_k$ as we will need to consider all $x_k \in \mX$, where we index $\mX$ using $k\in \mN$. Using this and $\delta_{k,t}$ indicator which is zero or one depending on whether action $k$ was played at time $t$. Note that, 
\begin{eqnarray*}
R_T & \leq & \sum_{t=1}^T (x)^\top \theta - x^\top \theta  = \sum_{t=1}^T (x)^\top \theta - x_t^\top\tilde{\theta}_t + x_t^\top\tilde{\theta}_t - x_t^\top \theta \\
& \stackrel{\text{def. UCB}} \leq & \sum_{t=1}^T x_t^\top (\tilde{\theta}_t - \theta) = \sum_{t=1}^T\sum_{k=1}^K \delta_{k,t} x_t^\top (\tilde{\theta}_t - \theta) \leq \sum_{t=1}^T\sum_{k=1}^K \delta_{k,t} |x_t^\top (\tilde{\theta}_t - \theta)|\\
& = & \sum_{t=1}^T\sum_{k=1}^K \delta_{k,t} |x_t^\top (\tilde{\theta}_t - \theta)\bW_{t,k}^{1/2}\bW_{t,k}^{-1/2}| = \sum_{t=1}^T\sum_{k=1}^K \delta_{k,t} |x_t^\top (\tilde{\theta}_t - \theta)\bW_{t,k}^{1/2}||\bW_{t,k}^{-1/2}| \\
& = & \sum_{t=1}^T\sum_{k=1}^K \delta_{k,t} \underbrace{\sqrt{x_k^\top \bV_t^{-1}x_k}}_{\bW_{t,k}^{-1/2}} \norm{x_k^\top(\tilde{\theta}_t - \theta)}_{\bW_{t,k}} \leq \sum_{t=1}^T\sum_{k=1}^K \delta_{k,t} \sqrt{x_k^\top \bV_t^{-1}x_k} \norm{x_k^\top(\tilde{\theta}_t - \theta)}_{\bOmega_{t,k}}  \\
& \stackrel{\text{Thm.}~\ref{thm:adapt_design_ridge_general}}\leq & \sum_{t=1}^T\sum_{k=1}^K \delta_{k,t}\sqrt{ x_k^\top \bV_t^{-1}x_k} \beta_{t,k} \\
& \stackrel{\text{Lemma~\ref{lemma:conf-param-size}}} \leq &\sum_{t=1}^T\sum_{k=1}^K \delta_{k,t} \norm{x_k}_{\bV_t^{-1}} \sqrt{\log(|\mX|T/\delta)} \\
& \stackrel{\text{Potential Lemma}} \leq & \sqrt{\log(|\mX|T/\delta)}  \sum_{t=1}^T \norm{x_t}_{\bV_t^{-1}} \leq \mO(\sqrt{dT\log((|\mX|+1)T/\delta)})
\end{eqnarray*}
which finishes the proof. The $\tilde{\theta}_t$ is the vector that corresponds to the UCB action played by the algorithm. We took the union bound for the adaptive confidence bounds for each $x \in \mX$, and hence the logarithmic dependence on $|\mX|$.
\end{proof}

\section{Matrix Algebra Results}

\begin{definition}[Matrix slices and weighted slices]
	
	\begin{eqnarray} \label{eq:smaller_splice}
		\bM_{SS} \eqdef \bI_{:S}^\top \bM \bI_{:S} & ~ \text{and} ~ & \bM_{S\eta>0} = \bD(\eta>0) \bM\bD(\eta>0)
	\end{eqnarray}	
	
	\begin{eqnarray}\label{eq:slice}
		\bM_S \eqdef \bI_{:S}\bM_{SS} \bI_{:S}^\top & ~ \text{and} ~ & \bM_{S\eta} = \bD(\eta) \bM\bD(\eta)
	\end{eqnarray}
	
\end{definition}

\begin{lemma}[\cite{Zhang2011}] \label{lemma:order}
	Let M be a positive definite matrix, and $S$ be a subset of $[n]$, then
	\begin{equation}\label{eq:inv}
		(\mathbf{M}_S)^{\dagger} \preceq (\mathbf{M}^{-1})_S
	\end{equation}
\end{lemma}

\begin{lemma}[\cite{Zhang2011}]\label{lemma:monotonicity}
	If $\bA \succeq \bB$ where $\bA,\bB \in \mR^{l\times l}$ then for any $X \in \mR^{d\times l}$, $\bX\bA\bX^\top \succeq \bX\bB\bX^\top$.
\end{lemma}

\begin{lemma} \label{lemma:projection} Let $ \bA \in \mR^{k \times m}$, then the matrix $\bP = \bA^\top (\bA\bA^\top )^{-1}\bA$ is projection matrix. 
\end{lemma}
\begin{proof}
	Its easy to check $\bP^\top =  \bP$ and $\bP^2 = \bP$. 
\end{proof}

\begin{lemma} Let $\bA \in \mR^{d\times d}$ full rank, and $\bC \in \mR^{k\times d}$ with rank $k$. Then $\bC \bA \bC^\top$ has rank $k$.
\end{lemma}

%

\begin{proposition} \label{prop:proposition-inv-pseudo} Let $\bX \in \mR^{m \times n}$ s.t. $\min(m,n) \geq |\eta|$, where $\bX$ is full rank,
	
	\[ \bV(\eta)^\dagger = \bX^\top \bD(\eta) ((\bX \bX^\top)_{S\eta})^\dagger \bD(\eta)  ((\bX \bX^\top)_{S\eta})^\dagger \bD(\eta) \bX \]
	Also, 
	\[	\Tr(\bV^{\dagger}) = \Tr(\bD(\eta)^{1/2} ((\bX \bX^\top)_{S\eta})^\dagger \bD(\eta)^{1/2}) \]
	
	If $n = |\eta|$, then 
	
	\begin{equation} \label{eq:proposition-inv-pseudo}
		\bV(\eta)^\dagger = \bX^\top (\bX \bX^\top)^{-1} \bD(\eta)^{-1}  (\bX \bX^\top)^{-1} \bX = \bX^\dagger \bD(\eta)^{-1} (\bX^\top)^\dagger.	
	\end{equation}
\end{proposition}

\begin{proof}
	Tedious verification of four pseudo-inverse criteria.
\end{proof}

\subsection{Auxiliary}

\begin{lemma}\label{lemma:generalized}
	Let $\bA \in \mR^{l\times l} \succ 0$, and $\bX \in \mR^{d\times l}$ diagonal where $d \leq l$, then 
	
	\begin{equation}
		\bA \succeq \bX^\top (\bX\bA^{-1}\bX^\top)^{-1}\bX.
	\end{equation}
\end{lemma}
\begin{proof}
	Let $S$ be a set that selects exactly the non-zero elements of matrix $\bX$ and further suppose w.l.o.g. the matrix is in such permutations that this block is right-upper most.
	
	By Lemma \ref{lemma:order}, we know that
	$ (\bA_{SS})^{-1} \preceq (\bA^{-1})_{SS} $. Due to the monotonicity of the operation (Lemma \ref{lemma:monotonicity})
	$\bX_{SS}(\bA_{SS})^{-1}\bX_{SS}^{\top} \preceq \bX_{SS}(\bA^{-1})_{SS}\bX_{SS}^{\top}$. Inverse operator reverse the above inequality, hence, 
	\begin{eqnarray}
		(\bX_{SS}(\bA_{SS})^{-1}\bX_{SS}^{\top})^{-1} & \succeq & (\bX_{SS}(\bA^{-1})_{SS}\bX_{SS}^{\top})^{-1} \\
		(\bX_{SS}^{-\top}(\bA_{SS})\bX_{SS}^{-1}) & \succeq & (\bX_{SS}(\bA^{-1})_{SS}\bX_{SS}^{\top})^{-1} \\
		(\bY_{SS}^{\top}(\bA_{SS})\bY_{SS}) & \succeq & (\bX_{SS}(\bA^{-1})_{SS}\bX_{SS}^{\top})^{-1}	\\
		(\bA_{SS}) & \succeq & \bX_{SS}^\top(\bX_{SS}(\bA^{-1})_{SS}\bX_{SS}^{\top})^{-1}\bX_{SS}
	\end{eqnarray}
	where definite following matrix $\bY$ s.t. $\bY_{SS} = (\bX_{SS})^{-1}$ and zero-otherwise. 
	
	Lastly, we know that $\bA_{S}$ has eigenvalues strictly smaller than $\bA$, and
	also note that $\bX \bA^{-1} \bX^\top =  \bX_{SS}(\bA^{-1})\bX_{SS}^{\top}$. 
	Thus applying Lemma \ref{lemma:monotonicity} we can lift the expression to $l\times l$ matrices, 
	\[	(\bA) \succeq \bI_{:S} (\bA)_{SS} \bI_{S:}\succeq \bI_{:S}\bX_{SS}^\top(\bX_{SS}(\bA^{-1})_{SS}\bX_{SS}^{\top})^{-1}\bX_{SS}\bI_{S:} \succeq  \bX^\top (\bX_{SS}(\bA^{-1})_{SS}\bX_{SS}^{\top})^{-1}\bX \]
	\[ =\bX (\bX \bA^{-1} \bX)^{-1}\bX \] 
\end{proof}

\begin{theorem}\label{thm:ranking}
	Let $\bA \in \mR^{l\times l} \succ 0$, and $\bM \in \mR^{d\times l}$, where $d \leq l$, and $\operatorname{rank}(\bM) = d$ then 
	
	\begin{equation}\label{eq:ranking}
		\bM^\top (\bM\bA^{-1}\bM^\top)^{-1}\bM \preceq \bA
	\end{equation}
\end{theorem}
\begin{proof}
	Let $\bM =\bU \bS \bV^\top$, where $\bU \in \mR^{d\times d}$ and $\bV \in \mR^{l\times l}$, be an SVD decomposition of $\bM$ and $\bA = \bR^\top\bLambda \bR$, where $\bR \in \mR^{l \times l}$. be eigendecomposition of $\bA$. 
	
	We perform the set of following equivalent operations to reduce the problem to diagonal case, 
	\begin{eqnarray}
		\bR^\top\bLambda \bR & \succeq & (\bU \bS \bV^\top)^\top(\bU \bS \bV^\top \bR^\top\bLambda^{-1} \bR (\bU \bS \bV^\top)^\top )^{-1}\bU \bS \bV^\top \\
		\bR^\top\bLambda \bR & \succeq& \bV \bS^\top \bU^\top(\bU \bS \bV^\top \bR^\top\bLambda^{-1} \bR \bV \bS^\top \bU )^{-1}\bU \bS \bV^\top \\
		\bR^\top\bLambda \bR & \succeq& \bV \bS^\top ( \bS \bV^\top \bR^\top\bLambda^{-1} \bR \bV \bS^\top  )^{-1} \bS \bV^\top\\
		\bLambda & \succeq& \bG \bS^\top ( \bS \bG^\top \bLambda^{-1} \bG \bS^\top  )^{-1} \bS \bG^{\top} \\
		\bB & \succeq& \bS^\top (\bS \bB^{-1} \bS^\top)^{-1} \bS 
	\end{eqnarray}
	where $\bV = \bR^\top \bG$ is a new rotation matrix, and $\bB = \bG^\top \bLambda \bG = \bV^\top \bR^\top \bLambda \bR\bV = \bV^\top \bA \bV$. The rest follows from Lemma \ref{lemma:generalized}.
\end{proof}

\begin{proposition}
	Let $\bX \in \mR^{n\times m}$, and $\bM \in \mR^{k\times m}$, where $k \leq m$, and $\operatorname{rank}(\bM) = k$ then if $\exists \bA$ s.t. $\bM = \bA \bX$, where $\bA \in {k \times n}$ and $k\leq n$, the following holds,
	
	\begin{equation}\label{eq:ranking2}
		\bM^\top ( \bM (\bX^\top \bX)^+ \bM^\top     )^{-1} \bM \preceq	 		\bX^\top \bX.
	\end{equation}
\end{proposition}

\begin{proof}
	The pseudo-inverse of $(\bX^\top \bX)^+ = \bX^\top (\bX\bX^\top)^{-2} \bX$. Consequently, 
	\begin{eqnarray}
		\bM^\top ( \bM (\bX^\top \bX)^+ \bM^\top     )^{-1} \bM & = &		\bX^\top \bA^\top ( \bA\bX \bX^\top (\bX\bX^\top)^{-2} \bX \bX^\top \bA^\top     )^{-1} \bA \bX\\
		& = & 	\bX^\top \bA^\top (\bA \bA^\top)^{-1}\bA \bX \preceq \bX^\top \bX
	\end{eqnarray}
	where the last line follows from the properties of a projection. 
\end{proof}
\end{document}